%% file: main.tex
\documentclass{article}
\usepackage{microtype}
\usepackage{graphicx}
\usepackage{caption}
\usepackage{subfigure}
\usepackage{amsmath}
\usepackage{amssymb}
\usepackage{mathtools}
\usepackage{amsthm}
\usepackage{amsfonts}
\usepackage{nicefrac}
\usepackage{float}
\usepackage{booktabs}
\usepackage[dvipsnames]{xcolor}
\usepackage{colortbl}
\usepackage{algorithm}
\usepackage[noend]{algorithmic}
\usepackage{bbding}
\usepackage{fancyvrb}
\usepackage{listings}
\definecolor{codegreen}{rgb}{0,0.5,0}
\definecolor{codered}{rgb}{0.7,0.1,0.1}
\definecolor{codegray}{rgb}{0.5,0.5,0.5}
\definecolor{codepurple}{rgb}{0.58,0,0.82}
\definecolor{backcolour}{rgb}{1,1,1}
\lstdefinestyle{python}{
    language=Python,
    backgroundcolor=\color{backcolour},   
    commentstyle=\color{codered}\textit,
    keywordstyle=\bfseries\color{codegreen},
    numberstyle=\tiny\color{codegray},
    stringstyle=\color{codepurple},
    basicstyle=\ttfamily\scriptsize,
    breakatwhitespace=false,         
    breaklines=true,                 
    captionpos=b,                    
    keepspaces=true,                 
    numbers=left,                    
    numbersep=4pt,                  
    showspaces=false,                
    showstringspaces=false,
    showtabs=false,                  
    tabsize=1,
    fancyvrb=true
}
\newenvironment{codesnippet}
  { \VerbatimEnvironment%
    \begin{Verbatim} }
  { \end{Verbatim}  }
\usepackage{hyperref}

\newcommand{\colorcell}{\cellcolor{CadetBlue!8}}
\newcommand{\hcolorcell}{\cellcolor{YellowGreen!10}}
\usepackage[accepted]{icml2022}
\definecolor{citecolor}{HTML}{0071bc}

\icmltitlerunning{TD-Learning for MPC}

\begin{document}

\twocolumn[
\icmltitle{Temporal Difference Learning for Model Predictive Control}

\icmlsetsymbol{equal}{*}

\begin{icmlauthorlist}
\icmlauthor{Nicklas Hansen}{ucsd}
\icmlauthor{Xiaolong Wang}{equal,ucsd}
\icmlauthor{Hao Su}{equal,ucsd}
\end{icmlauthorlist}

\icmlaffiliation{ucsd}{UC San Diego}

\icmlcorrespondingauthor{Nicklas Hansen}{nihansen@ucsd.edu}

\icmlkeywords{Machine Learning, ICML, Reinforcement Learning, TD-Learning, Model Predictive Control}

\vskip 0.3in
]

\printAffiliationsAndNotice{\icmlEqualContribution}

\begin{abstract}
Data-driven model predictive control has two key advantages over model-free methods: a potential for improved sample efficiency through model learning, and better performance as computational budget for planning increases. However, it is both costly to plan over long horizons and challenging to obtain an accurate model of the environment. In this work, we combine the strengths of model-free and model-based methods. We use a learned task-oriented latent dynamics model for local trajectory optimization over a short horizon, and use a learned terminal value function to estimate long-term return, both of which are \emph{learned jointly} by temporal difference learning. Our method, \textbf{TD-MPC}, achieves superior sample efficiency and asymptotic performance over prior work on both state and image-based continuous control tasks from DMControl and Meta-World. Code and videos are available at \url{https://nicklashansen.github.io/td-mpc}.
\end{abstract}

\section{Introduction}
\label{sec:introduction}
To achieve desired behavior in an environment, a Reinforcement Learning (RL) agent needs to iteratively interact and consolidate knowledge about the environment. Planning is a powerful approach to such sequential decision making problems, and has achieved tremendous success in application areas such as game-playing \cite{Kaiser2020ModelBasedRL, Schrittwieser2020MasteringAG} and continuous control \cite{Tassa2012SynthesisAS, Chua2018DeepRL, Janner2019WhenTT}. By utilizing an internal model of the environment, an agent can plan a trajectory of actions ahead of time that leads to the desired behavior; this is in contrast to \textit{model-free} algorithms that learn a policy purely through trial-and-error.

\begin{figure}[t!]
    \centering
    \vspace{-0.1in}
    \includegraphics[width=0.4\textwidth]{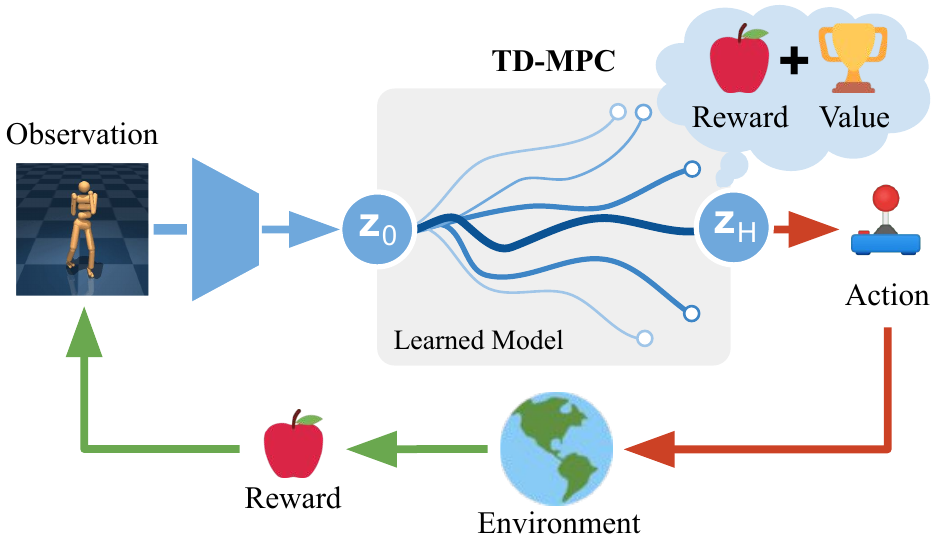}\vspace{0.075in}\\
    \includegraphics[width=0.48\textwidth]{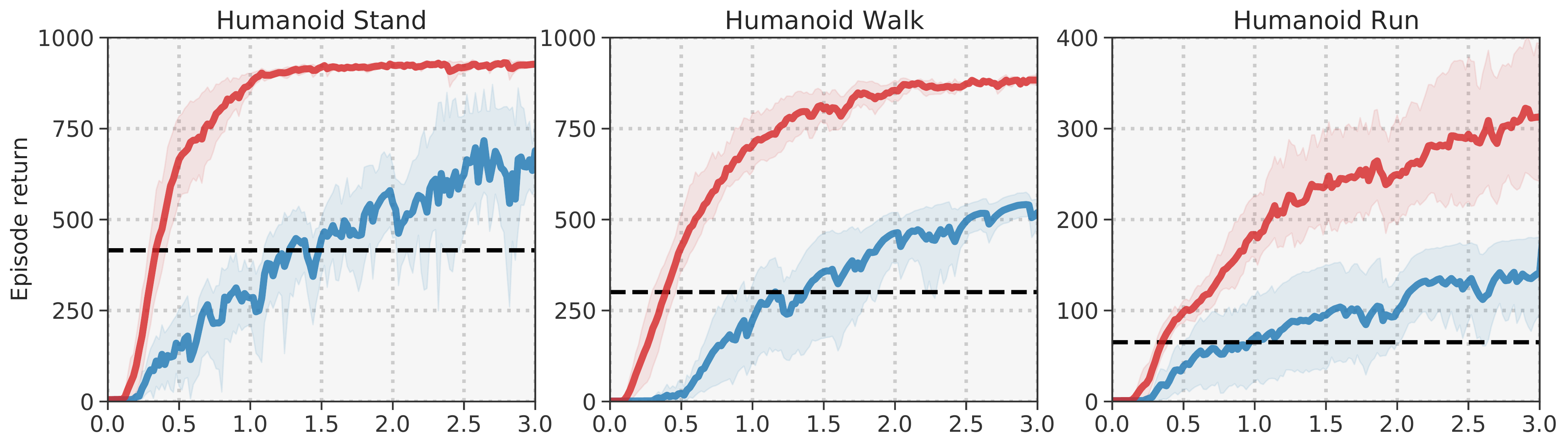}\vspace{0.025in}\\
    \includegraphics[width=0.48\textwidth]{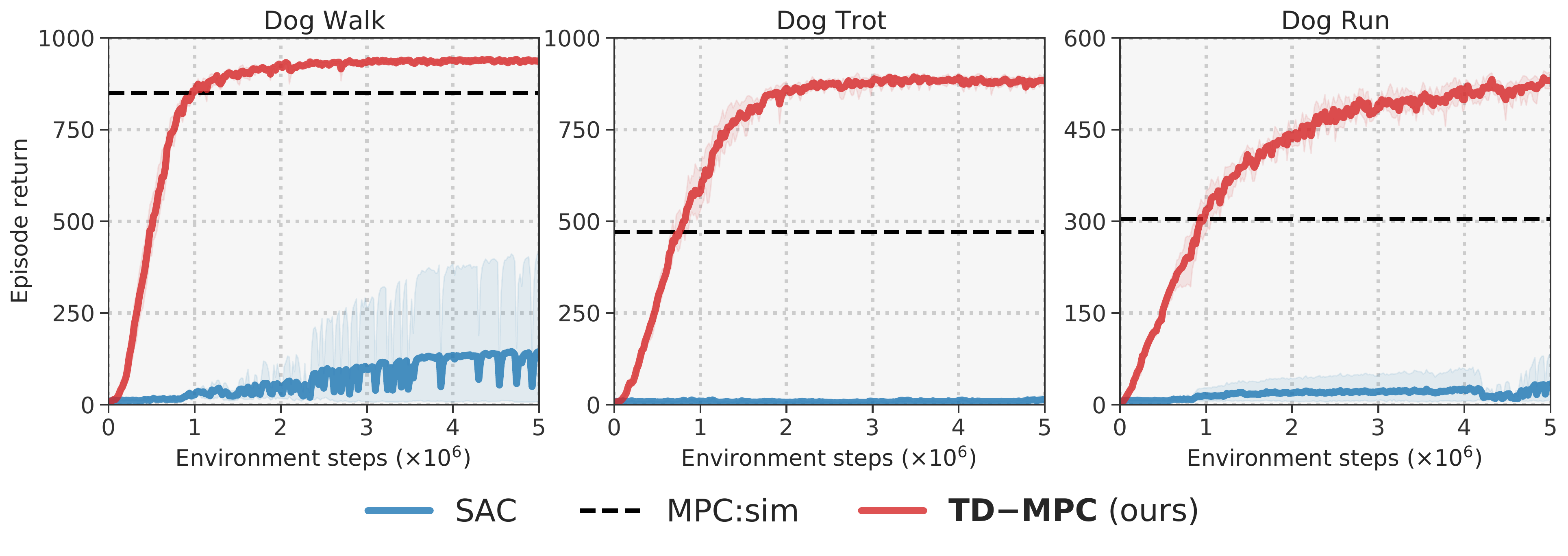}
    \vspace{-0.325in}
    \caption{\textbf{Overview.} \textit{(Top)} We present a framework for MPC using a task-oriented latent dynamics model and value function learned jointly by temporal difference learning. We perform trajectory optimization over model rollouts and use the value function for long-term return estimates. \textit{(Bottom)} Episode return of our method, SAC, and MPC with a ground-truth simulator on challenging, high-dimensional Humanoid and Dog tasks \citep{deepmindcontrolsuite2018}. Mean of 5 runs; shaded areas are $95\%$ confidence intervals.}
    \label{fig:planning}
    \vspace{-0.2in}
\end{figure}

Concretely, prior work on model-based methods can largely be subdivided into two directions, each exploiting key advantages of model-based learning: \textit{(i)} planning, which is advantageous over a learned policy, but it can be prohibitively expensive to plan over long horizons \citep{Janner2019WhenTT, Lowrey2019PlanOL, hafner2019planet, Argenson2021ModelBasedOP}; and \textit{(ii)} using a learned model to improve sample-efficiency of model-free methods by e.g. learning from generated rollouts, but this makes model biases likely to propagate to the policy as well \citep{ha2018worldmodels, Hafner2020DreamTC, Clavera2020ModelAugmentedAB}. As a result, model-based methods have historically struggled to outperform simpler, model-free methods \citep{srinivas2020curl, kostrikov2020image} in continuous control tasks.

Can we instead augment model-based planning with the strengths of model-free learning? Because of the immense cost of long-horizon planning, Model Predictive Control (MPC) optimizes a trajectory over a shorter, finite horizon, which yields only temporally local optimal solutions. MPC can be extended to approximate globally optimal solutions by using a terminal value function that estimates discounted return beyond the planning horizon. However, obtaining an accurate model and value function can be challenging.

In this work, we propose \textbf{T}emporal \textbf{D}ifference Learning for \textbf{M}odel \textbf{P}redictive \textbf{C}ontrol (\textbf{TD-MPC}), a framework for data-driven MPC using a task-oriented latent dynamics model and terminal value function \emph{learned jointly} by temporal difference (TD) learning. At each decision step, we perform trajectory optimization using short-term reward estimates generated by the learned model, and use the learned value function for long-term return estimates. For example, in the Humanoid locomotion task shown in Figure \ref{fig:planning}, planning with a model may be beneficial for accurate joint movement, whereas the higher-level objective, e.g. direction of running, can be guided by long-term value estimates.

A key technical contribution is how the model is learned. While prior work learns a model through state or video prediction, we argue that it is remarkably inefficient to model everything in the environment, including irrelevant quantities and visuals such as shading, as this approach suffers from model inaccuracies and compounding errors. To overcome these challenges, we make three key changes to model learning. Firstly, we learn the latent representation of the dynamics model purely from rewards, ignoring nuances unnecessary for the task at hand. This makes the learning more sample efficient than state/image prediction. Secondly, we back-propagate gradients from the reward and TD-objective through multiple rollout steps of the model, improving reward and value predictions over long horizons. This alleviates error compounding when conducting rollouts. Lastly, we propose a modality-agnostic prediction loss \emph{in latent space} that enforces temporal consistency in the learned representation \emph{without} explicit state or image prediction.

We evaluate our method on a variety of continuous control tasks from DMControl \citep{deepmindcontrolsuite2018} and Meta-World \citep{yu2019meta}, where we find that our method achieves superior sample efficiency and asymptotic performance over prior model-based and model-free methods. In particular, our method solves Humanoid and Dog locomotion tasks with up to 38-dimensional continuous action spaces in as little as 1M environment steps (see Figure \ref{fig:planning}), and is trivially extended to match the state-of-the-art in image-based RL.

\section{Preliminaries}
\label{sec:preliminaries}
\textbf{Problem formulation.} We consider infinite-horizon Markov Decision Processes (MDP) characterized by a tuple $(\mathcal{S}, \mathcal{A}, \mathcal{T}, \mathcal{R}, \gamma, p_{0})$, where $\mathcal{S} \in \mathbb{R}^{n}$ and $\mathcal{A} \in \mathbb{R}^{m}$ are continuous state and action spaces, $\mathcal{T} \colon \mathcal{S} \times \mathcal{A} \times \mathcal{S} \mapsto \mathbb{R}_{+}$ is the transition (dynamics) function, $\mathcal{R} \colon \mathcal{S} \times \mathcal{A} \mapsto \mathbb{R}$ is a reward function, $\gamma \in [0,1)$ is a discount factor, and $p_{0}$ is the initial state distribution. We aim to learn a parameterized mapping $\Pi_{\theta} \colon \mathcal{S} \mapsto \mathcal{A}$ with parameters $\theta$ such that discounted return $\mathbb{E}_{\Gamma \sim \Pi_{\theta}} [ \sum^{\infty}_{t=1} \gamma^{t} r_{t} ],~r_{t} \sim \mathcal{R}(\cdot | \mathbf{s}_{t}, \mathbf{a}_{t})$ is maximized along a trajectory $\Gamma = (\mathbf{s}_{0}, \mathbf{a}_{0}, \mathbf{s}_{1}, \mathbf{a}_{1},\dots)$ following $\Pi_{\theta}$ by sampling an action $\mathbf{a}_{t} \sim \Pi_{\theta}(\cdot | \mathbf{s}_{t})$ and reaching state $\mathbf{s}_{t+1} \sim \mathcal{T}(\cdot | \mathbf{s}_{t}, \mathbf{a}_{t})$ at each decision step $t$.

\textbf{Fitted $Q$-iteration.} Model-free TD-learning algorithms aim to estimate an optimal state-action value function $Q^{*} \colon \mathcal{S} \times \mathcal{A} \mapsto \mathbb{R}$ using a parametric value function $Q_{\theta}(\mathbf{s}, \mathbf{a}) \approx Q^{*}(\mathbf{s}, \mathbf{a}) = \max_{\mathbf{a}'} \mathbb{E}[ \mathcal{R}(\mathbf{s}, \mathbf{a}) + \gamma Q^{*}(\mathbf{s}', \mathbf{a}')]~\forall \mathbf{s} \in \mathcal{S}$ where $\mathbf{s}', \mathbf{a}'$ is the state and action at the following step, and $\theta$ parameterizes the function \citep{Sutton1988LearningTP}. For $\gamma \approx 1$, $Q^{*}$ estimates discounted return for the optimal policy over an infinite horizon. While $Q^{*}$ is generally unknown, it can be approximated by repeatedly fitting $Q_{\theta}$ using the update rule
\begin{align}
    \label{eq:fitted-value-iteration}
    \theta^{k+1} \leftarrow \arg\min_{\theta} \mathbb{E}_{(\mathbf{s}, \mathbf{a}, \mathbf{s}') \sim \mathcal{B}}~\| Q_{\theta}(\mathbf{s}, \mathbf{a}) - y \|_{2}^{2}
\end{align}
where the $Q$-\textit{target} $y =  \mathcal{R}(\mathbf{s}, \mathbf{a}) + \gamma \max_{\mathbf{a}'} Q_{\theta^{-}}(\mathbf{s}', \mathbf{a}')$, $\mathcal{B}$ is a replay buffer that is iteratively grown as new data is collected, and $\theta^{-}$ is a slow-moving average of the online parameters $\theta$ updated with the rule $\theta^{-}_{k+1} \longleftarrow (1 - \zeta) \theta^{-}_{k} + \zeta \theta_{k}
$ at each iteration using a constant coefficient $\zeta \in [0, 1)$.

\textbf{Model Predictive Control.} In actor-critic RL algorithms, $\Pi$ is typically a policy parameterized by a neural network that learns to approximate $\Pi_{\theta}(\cdot | \mathbf{s}) \approx \arg\max_{\mathbf{a}} \mathbb{E}[ Q_{\theta}(\mathbf{s}, \mathbf{a}) ]~\forall \mathbf{s} \in \mathcal{S}$, i.e, the globally optimal policy. In control, $\Pi$ is traditionally implemented as a trajectory optimization procedure. To make the problem tractable, one typically obtains a \textit{local} solution to the trajectory optimization problem at each step $t$ by estimating optimal actions $\mathbf{a}_{t:t+H}$ over a finite horizon $H$ and executing the first action $\mathbf{a}_{t}$, known as \textit{Model Predictive Control} (MPC):
\begin{equation}
    \label{eq:mpc}
    \Pi^{\text{MPC}}_{\theta}(\mathbf{s}_{t}) = \arg\max_{\mathbf{a}_{t:t+H}} \mathbb{E}\left[ \sum_{i=t}^{H} \gamma^{i} \mathcal{R}(\mathbf{s}_{i}, \mathbf{a}_{i}) \right]\,,
\end{equation}
where $\gamma$, unlike in fitted $Q$-iteration, is typically set to 1, i.e., no discounting. Intuitively, Equation \ref{eq:mpc} can be viewed as a special case of the standard additive-cost optimal control objective. A solution can be found by iteratively fitting parameters of a family of distributions, e.g., $\mu, \sigma$ for a multivariate Gaussian with diagonal covariance, to the space of actions over a finite horizon using the derivative-free Cross-Entropy Method (CEM; \citet{Rubinstein1997OptimizationOC}), and sample trajectories generated by a model. As opposed to fitted $Q$-iteration, Equation \ref{eq:mpc} is not predictive of long-term rewards, hence a \textit{myopic} solution. When a value function is known (e.g. a heuristic or in the context of our method: estimated using Equation \ref{eq:fitted-value-iteration}), it can be used in conjunction with Equation \ref{eq:mpc} to estimate discounted return at state $\mathbf{s}_{t+H}$ and beyond; such methods are known as MPC with a \textit{terminal value function}. In the following, we consider parameterized mappings $\Pi$ from both the perspective of actor-critic RL algorithms and model predictive control (planning). To disambiguate these concepts, we refer to planning with MPC as $\Pi_{\theta}$ and a policy network as $\pi_{\theta}$. We generically denote parameterization using neural networks as $\theta$ (online) and $\theta^{-}$ (\textit{target}; slow-moving average of $\theta$) as combined feature vectors.

\section{TD-Learning for Model Predictive Control}
\label{sec:planning}
\vspace{-0.025in}
We propose \textbf{TD-MPC}, a framework that combines MPC with a task-oriented latent dynamics model and terminal value function jointly learned using TD-learning in an online RL setting. Specifically, TD-MPC leverages Model Predictive Path Integral (MPPI; \citet{Williams2015ModelPP}) control for planning (denoted $\Pi_{\theta}$), learned models $d_{\theta}, R_{\theta}$ of the (latent) dynamics and reward signal, respectively, a terminal state-action value function $Q_{\theta}$, and a parameterized policy $\pi_{\theta}$ that helps guide planning. We summarize our framework in Figure \ref{fig:planning} and Algorithm~\ref{alg:inference}. In this section, we detail the inference-time behavior of our method, while we defer discussion of training to Section \ref{sec:latent-dynamics-model}.

MPPI is an MPC algorithm that iteratively updates parameters for a family of distributions using an importance weighted average of the estimated top-$k$ sampled trajectories (in terms of expected return); in practice, we fit parameters of a time-dependent multivariate Gaussian with diagonal covariance. We adapt MPPI as follows. Starting from initial parameters $(\mu^{0}, \sigma^{0})_{t:t+H},~\mu^{0},\sigma^{0} \in \mathbb{R}^{m},~\mathcal{A} \in \mathbb{R}^{m}$, i.e. independent parameters for each action over a horizon of length $H$, we independently sample $N$ trajectories using rollouts generated by the learned model $d_{\theta}$, and estimate the total return $\phi_{\Gamma}$ of a sampled trajectory $\Gamma$ as
\begin{equation}
    \label{eq:mppi-return}
    \phi_{\Gamma} \triangleq \mathbb{E}_{\Gamma}\left[ \gamma^{H} Q_{\theta}(\mathbf{z}_{H}, \mathbf{a}_{H}) + \sum_{t=0}^{H-1} \gamma^{t} R_{\theta}(\mathbf{z}_{t}, \mathbf{a}_{t}) \right]\,,
\end{equation}
where $\mathbf{z_{t+1}} = d_{\theta}(\mathbf{z}_{t}, \mathbf{a}_{t})$ and $\mathbf{a}_{t} \sim \mathcal{N}(\mu^{j-1}_{t}, (\sigma^{j-1}_{t})^{2} \mathrm{I})$ at iteration $j-1$, as highlighted in {\color{BrickRed}red} in Algorithm \ref{alg:inference}. We select the top-$k$ returns $\phi_{\Gamma}^{\star}$ and obtain new parameters $\mu^{j},\sigma^{j}$ at iteration $j$ from a $\phi_{\Gamma}^{\star}$-normalized empirical estimate:
\begin{align}
    \label{eq:mppi-return-norm}
    &\mu^{j} = \frac{\sum_{i=1}^{k} \Omega_{i} \Gamma_{i}^{\star}}{\sum_{i=1}^{k} \Omega_{i}}\,,~
    \sigma^{j} = \sqrt{\frac{\sum_{i=1}^{k} \Omega_{i} (\Gamma_{i}^{\star} - \mu^{j})^{2}}{\sum_{i=1}^{k} \Omega_{i}}}\,,
\end{align}
where $\Omega_{i} = e^{\tau (\phi_{\Gamma,i}^{\star})}$, $\tau$ is a temperature parameter controlling the ``sharpness" of the weighting, and $\Gamma_{i}^{\star}$ denotes the $i$th top-$k$ trajectory corresponding to return estimate $\phi_{\Gamma}^{\star}$. After a fixed number of iterations $J$, the planning procedure is terminated and a trajectory is sampled from the final return-normalized distribution over action sequences. We plan at each decision step $t$ and execute only the first action, i.e., we employ \emph{receding-horizon} MPC to produce a feedback policy. To reduce the number of iterations required for convergence, we ``warm start" trajectory optimization at each step $t$ by reusing the 1-step shifted mean $\mu$ obtained at the previous step \citep{Argenson2021ModelBasedOP}, but always use a large initial variance to avoid local minima.

\begin{figure}[t]
\vspace{-0.15in}
\begin{algorithm}[H]
\caption{~~TD-MPC {\color{CadetBlue}(\emph{inference})}}
\label{alg:inference}
\begin{algorithmic}[1]
\REQUIRE $\theta:$ learned network parameters\\
~~~~~~~~~~$\mu^{0}, \sigma^{0}$: initial parameters for $\mathcal{N}$\\
~~~~~~~~~~$N, N_{\pi}$: num sample/policy trajectories\\
~~~~~~~~~~$\mathbf{s}_{t}, H$: current state, rollout horizon
\STATE Encode state $\mathbf{z}_{t} \leftarrow h_{\theta}(\mathbf{s}_{t})$~~~~~~~$\vartriangleleft$ {\color{CadetBlue}\emph{Assuming TOLD model}}
\FOR{each iteration $j=1..J$}
\STATE Sample $N$ traj. of len. $H$ from $\mathcal{N}(\mu^{j-1}, (\sigma^{j-1})^{2} \mathrm{I})$
\STATE {\color{MidnightBlue} Sample $N_{\pi}$ traj. of length $H$ using $\pi_{\theta}, d_{\theta}$}\\ {\color{CadetBlue} \emph{// Estimate trajectory returns $\phi_{\Gamma}$ using $d_{\theta},R_{\theta},Q_{\theta}$,\\~~~starting from $\mathbf{z}_{t}$ and initially letting $\phi_{\Gamma}=0$:}}
\FOR{all $N+N_{\pi}$ trajectories $(\mathbf{a}_{t}, \mathbf{a}_{t+1},\dots, \mathbf{a}_{t+H})$}
{\color{BrickRed}
\FOR{step $t=0..H-1$}
\STATE $\phi_{\Gamma} = \phi_{\Gamma} + \gamma^{t} R_{\theta}(\mathbf{z}_{t}, \mathbf{a}_{t})$~~~~~~~~~~~~~~~~~~~{\color{CadetBlue}$\vartriangleleft$ \emph{Reward}}
\STATE $\mathbf{z}_{t+1} \leftarrow d_{\theta}(\mathbf{z}_{t}, \mathbf{a}_{t})$~~~~~~~~~~~~~~~~{\color{CadetBlue}$\vartriangleleft$ \emph{Latent transition}}
\ENDFOR
\STATE $\phi_{\Gamma} = \phi_{\Gamma} + \gamma^{H} Q_{\theta}(\mathbf{z}_{H}, \mathbf{a}_{H})$~~~~~~~~{\color{CadetBlue}$\vartriangleleft$ \emph{Terminal value}}}
\ENDFOR
{\color{CadetBlue} \emph{// Update parameters $\mu,\sigma$ for next iteration:}}
\STATE $\mu^{j}, \sigma^{j} =$ Equation \ref{eq:mppi-return-norm} (and Equation \ref{eq:mppi-exploration})
\ENDFOR
\STATE \textbf{return} $\mathbf{a} \sim \mathcal{N}(\mu^{J}, (\sigma^{J})^{2} \mathrm{I})$
\end{algorithmic}
\end{algorithm}
\vspace{-0.3in}
\end{figure}

\textbf{Exploration by planning.} Model-free RL algorithms such as DDPG \citep{Lillicrap2016ContinuousCW} encourage exploration by injecting action noise (e.g. Gaussian or Ornstein-Uhlenbeck noise) into the learned policy $\pi_{\theta}$ during training, optionally following a linear annealing schedule. While our trajectory optimization procedure is inherently stochastic due to trajectory sampling, we find that the rate at which $\sigma$ decays varies wildly between tasks, leading to (potentially poor) local optima for small $\sigma$. To promote consistent exploration across tasks, we constrain the std. deviation of the sampling distribution such that, for a $\mu^{j}$ obtained from Equation \ref{eq:mppi-return-norm} at iteration $j$, we instead update $\sigma^{j}$ to
\begin{equation}
    \label{eq:mppi-exploration}
    \sigma^{j} = \max\left(\sqrt{\frac{\sum_{i=1}^{N} \Omega_{i} (\Gamma_{i}^{\star} - \mu^{j})^{2}}{\sum_{i=1}^{N} \Omega_{i}}}\,,~\epsilon\right)\,,
\end{equation}
where $\epsilon \in \mathbb{R}_{+}$ is a linearly decayed constant. Likewise, we linearly increase the planning horizon from $1$ to $H$ in the early stages of training, as the model is initially inaccurate and planning would therefore be dominated by model bias.

\textbf{Policy-guided trajectory optimization.} Analogous to \citet{Schrittwieser2020MasteringAG, Sikchi2020LearningOW}, TD-MPC learns a policy $\pi_{\theta}$ in addition to planning procedure $\Pi_{\theta}$, and augments the sampling procedure with additional samples from $\pi_{\theta}$ (highlighted in {\color{MidnightBlue}blue} in Algorithm \ref{alg:inference}). This leads to one of two cases: the policy trajectory is estimated to be \textit{(i)} poor, and may be \emph{excluded} from the top-$k$ trajectories; or \textit{(ii)} good, and may be \emph{included} with influence proportional to its estimated return $\phi_{\Gamma}$. While LOOP relies on the maximum entropy objective of SAC \citep{Haarnoja2018SoftAA} for exploration, TD-MPC learns a deterministic policy. To make sampling stochastic, we apply linearly annealed (Gaussian) noise to $\pi_{\theta}$ actions as in DDPG \citep{Lillicrap2016ContinuousCW}. Our full procedure is summarized in Algorithm \ref{alg:inference}.

\vspace{-0.05in}
\section{Task-Oriented Latent Dynamics Model}
\label{sec:latent-dynamics-model}
\begin{figure}[t]
    \centering
    \includegraphics[width=0.42\textwidth]{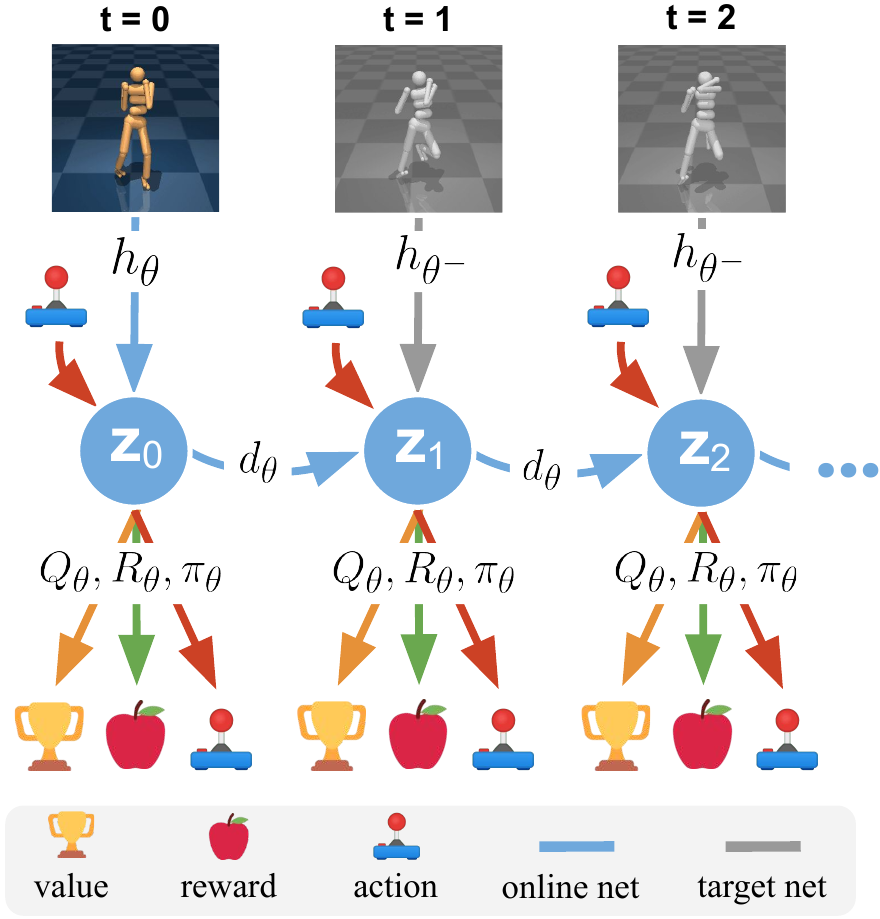}
    \vspace{-0.125in}
    \caption{\textbf{Training our TOLD model.} A trajectory $\Gamma_{0:H}$ of length $H$ is sampled from a replay buffer, and the first observation $\mathbf{s}_{0}$ is encoded by $h_{\theta}$ into a latent representation $\mathbf{z}_{0}$. Then, TOLD recurrently predicts the following latent states $\mathbf{z}_{1}, \mathbf{z}_{2},\dots,\mathbf{z}_{H}$, as well as a value $\hat{q}$, reward $\hat{r}$, and action $\hat{\mathbf{a}}$ for each latent state, and we optimize TOLD using Equation \ref{eq:objective}. Subsequent observations are encoded using target net $h_{\theta^{-}}$ ($\theta^{-}$: slow-moving average of $\theta$) and used as latent targets only during training (illustrated in gray).}
    \label{fig:architectural-overview}
    \vspace{-0.15in}
\end{figure}

\vspace{-0.05in}
To be used in conjunction with TD-MPC, we propose a \textbf{T}ask-\textbf{O}riented \textbf{L}atent \textbf{D}ynamics (TOLD) model that is jointly learned together with a terminal value function using TD-learning. Rather than attempting to model the environment itself, our TOLD model learns to only model elements of the environment that are predictive of reward, which is a far easier problem. During inference, our TD-MPC framework leverages the learned TOLD model for trajectory optimization, estimating short-term rewards using model rollouts and long-term returns using the terminal value function. TD-MPC and TOLD support continuous action spaces, arbitrary input modalities, and sparse reward signals. Figure \ref{fig:architectural-overview} provides an overview of the TOLD training procedure.

\textbf{Components.} Throughout training, our agent iteratively performs the following two operations: \textit{(i)} improving the learned TOLD model using data collected from previous environment interaction; and \textit{(ii)} collecting new data from the environment by online planning of action sequences with TD-MPC, using TOLD for generating imagined rollouts. Our proposed TOLD consists of five learned components $h_{\theta},d_{\theta},R_{\theta},Q_{\theta},\pi_{\theta}$ that predict the following quantities:
\begin{equation}
\begin{array}{lll}
    \text{Representation:} && \mathbf{z}_{t} = h_{\theta}(\mathbf{s}_{t})\\
    \text{Latent dynamics:} && \mathbf{z}_{t+1} = d_{\theta}(\mathbf{z}_{t}, \mathbf{a}_{t})\\
    \text{Reward:} && \hat{r}_{t} = R_{\theta}(\mathbf{z}_{t}, \mathbf{a}_{t})\\
    \text{Value:} && \hat{q}_{t} = Q_{\theta}(\mathbf{z}_{t}, \mathbf{a}_{t})\\
    \text{Policy:} && \hat{\mathbf{a}}_{t} \sim \pi_{\theta}(\mathbf{z}_{t})
\end{array}
\end{equation}
Given an observation $\mathbf{s}_{t}$ observed at time $t$, a representation network $h_{\theta}$ encodes $\mathbf{s}_{t}$ into a latent representation $\mathbf{z}_{t}$. From $\mathbf{z}_{t}$ and an action $\mathbf{a}_{t}$ taken at time $t$, TOLD then predicts \textit{(i)} the latent dynamics (latent representation $\mathbf{z}_{t+1}$ of the following timestep); \textit{(ii)} the single-step reward received; \textit{(iii)} its state-action ($Q$) value; and \textit{(iv)} an action that (approximately) maximizes the $Q$-function. To make TOLD less susceptible to compounding errors, we recurrently predict the aforementioned quantities multiple steps into the future from predicted future latent states, and back-propagate gradients through time. Unlike prior work \citep{ha2018worldmodels, Janner2019WhenTT, hafner2019planet, Hafner2020DreamTC, Sikchi2020LearningOW}, we find it sufficient to implement all components of TOLD as purely deterministic MLPs, i.e., \textit{without} RNN gating mechanisms nor probabilistic models.

\textbf{Objective.} We first state the full objective, and then motivate each module and associated objective term. During training, we minimize a temporally weighted objective
\begin{equation}
\label{eq:objective}
    \mathcal{J}(\theta; \Gamma) = \sum_{i=t}^{t+H} \lambda^{i-t} \mathcal{L}(\theta; \Gamma_{i})\,,
\end{equation}
where $\Gamma \sim \mathcal{B}$ is a trajectory $(\mathbf{s}_{t}, \mathbf{a}_{t}, r_{t}, \mathbf{s}_{t+1})_{t:t+H}$ sampled from a replay buffer $\mathcal{B}$, $\lambda \in \mathbb{R}_{+}$ is a constant that weights near-term predictions higher, and the single-step loss
\begin{align}
    \label{eq:reward-loss} &\mathcal{L}(\theta; \Gamma_{i}) = c_{1} {\color{CadetBlue}\underbrace{{\color{black} \| R_{\theta}(\mathbf{z}_{i}, \mathbf{a}_{i}) - r_{i} \|^{2}_{2}}}_\text{reward}}\\
    \label{eq:value-loss} &+ c_{2} {\color{CadetBlue}\underbrace{{\color{black} \| Q_{\theta}(\mathbf{z}_{i}, \mathbf{a}_{i}) - \left( r_{i} + \gamma Q_{\theta^{-}}(\mathbf{z}_{i+1}, \pi_{\theta}(\mathbf{z}_{i+1})) \right) \|^{2}_{2}}}_\text{value}}\\
    \label{eq:consistency-loss} &+ c_{3} {\color{CadetBlue}\underbrace{{\color{black}\| d_{\theta}(\mathbf{z}_{i}, \mathbf{a}_{i}) - h_{\theta^{-}}(\mathbf{s}_{i+1}) \|^{2}_{2}}}_{\text{latent state consistency}}}
\end{align}
is employed to jointly optimize for reward prediction, value prediction, and a latent state consistency loss that regularizes the learned representation. Here, $c_{1:3}$ are constant coefficients balancing the three losses. From each transition $(\mathbf{z}_{i}, \mathbf{a}_{i})$, the reward term (Equation \ref{eq:reward-loss}) predicts the single-step reward, the value term (Equation \ref{eq:value-loss}) is our adoption of fitted $Q$-iteration from Equation \ref{eq:fitted-value-iteration} following previous work on actor-critic algorithms \citep{Lillicrap2016ContinuousCW, Haarnoja2018SoftAA}, and the consistency term (Equation \ref{eq:consistency-loss}) predicts the latent representation of future states. Crucially, recurrent predictions are made entirely in latent space from states $\mathbf{z}_{i} = h_{\theta}(\mathbf{s}_{i})$, $\mathbf{z}_{i+1} = d_{\theta}(\mathbf{z}_{i}, \mathbf{a}_{i})$, $\dots$, $\mathbf{z}_{i+H} = d_{\theta}(\mathbf{z}_{i+H-1}, \mathbf{a}_{i+H-1})$ such that only the first observation $\mathbf{s}_{i}$ is encoded using $h_{\theta}$ and gradients \emph{from all three terms} are back-propagated \textit{through time}. This is in contrast to prior work on model-based learning that learn a model by state or video prediction, entirely decoupled from policy and/or value learning \citep{ha2018worldmodels, Hafner2020DreamTC, Sikchi2020LearningOW}. We use an exponential moving average $\theta^{-}$ of the online network parameters $\theta$ for computing the value target \cite{Lillicrap2016ContinuousCW}, and similarly also use $\theta^{-}$ for the latent state consistency target $h_{\theta^{-}}(\mathbf{s}_{i+1})$. The policy $\pi_{\theta}$ is described next, while we defer discussion of the consistency loss to the following section.

\textbf{Computing TD-targets.} The TD-objective in Equation \ref{eq:value-loss} requires estimating the quantity $\max_{\mathbf{a}_{t}} Q_{\theta^{-}}(\mathbf{z}_{t}, \mathbf{a}_{t})$, which is extremely costly to compute using planning \citep{Lowrey2019PlanOL}. Therefore, we instead learn a policy $\pi_{\theta}$ that maximizes $Q_{\theta}$ by minimizing the objective
\begin{equation}
    \label{eq:policy-objective}
    \mathcal{J}_{\pi}(\theta; \Gamma) = - \sum_{i=t}^{t+H} \lambda^{i-t} Q_{\theta}(\mathbf{z}_{i}, \pi_{\theta}(\operatorname{sg}(\mathbf{z}_{i})))\,,
\end{equation}
which is a temporally weighted adaptation of the policy objective commonly used in model-free actor-critic methods such as DDPG \cite{Lillicrap2016ContinuousCW} and SAC \cite{Haarnoja2018SoftAA}. Here, $\operatorname{sg}$ denotes the stop-grad operator, and Equation \ref{eq:policy-objective} is optimized only wrt. policy parameters. While we empirically observe that for complex tasks the learned $\pi_{\theta}$ is inferior to planning (discussed in Section \ref{sec:experiments}), we find it sufficiently expressive for efficient value learning.

\textbf{Latent state consistency.} To provide a rich learning signal for model learning, prior work on model-based RL commonly learn to directly predict future states or pixels \citep{ha2018worldmodels, Janner2019WhenTT, Lowrey2019PlanOL, Kaiser2020ModelBasedRL, Sikchi2020LearningOW}. However, learning to predict future observations is an extremely hard problem as it forces the network to model everything in the environment, including task-irrelevant quantities and details such as shading.
Instead, we propose to regularize TOLD with a \textit{latent state consistency loss} (shown in Equation \ref{eq:consistency-loss}) that forces a future latent state prediction $\mathbf{z}_{t+1} = d_{\theta}(\mathbf{z}_{t}, \mathbf{a}_{t})$ at time $t+1$ to be similar to the latent representation of the corresponding ground-truth observation $h_{\theta^{-}}(\mathbf{s}_{t+1})$, circumventing prediction of observations altogether. Additionally, this design choice effectively makes model learning agnostic to the observation modality. The training procedure is shown in Algorithm \ref{alg:training}; see Appendix \ref{sec:appendix-implementation-details} for pseudo-code.

\begin{figure}[t]
\vspace{-0.1in}
\begin{algorithm}[H]
\caption{~~TOLD {\color{CadetBlue}(\emph{training})}}
\label{alg:training}
\begin{algorithmic}[1]
\REQUIRE $\theta, \theta^{-}$: randomly initialized network parameters\\
~~~~~~~~~~$\eta, \tau, \lambda, \mathcal{B}$: learning rate, coefficients, buffer
\WHILE{not tired}
\STATE {\color{CadetBlue} \emph{// Collect episode with TD-MPC from $\mathbf{s}_{0} \sim p_{0}$:}}
\FOR{step $t=0...T$}
\STATE $\mathbf{a}_{t} \sim \Pi_{\theta}(\cdot | h_{\theta}(\mathbf{s}_{t}))$~~~~~~~~~~{\color{CadetBlue}$\vartriangleleft$ \emph{Sample with TD-MPC}}
\STATE $(\mathbf{s}_{t+1},~r_{t}) \sim \mathcal{T}(\cdot | \mathbf{s}_{t}, \mathbf{a}_{t}),~ \mathcal{R}(\cdot | \mathbf{s}_{t}, \mathbf{a}_{t})$~~~{\color{CadetBlue}$\vartriangleleft$ \emph{Step env.}}
\STATE $\mathcal{B} \leftarrow \mathcal{B} \cup (\mathbf{s}_{t}, \mathbf{a}_{t}, r_{t}, \mathbf{s}_{t+1})$~~~~~~~~~~~~{\color{CadetBlue}$\vartriangleleft$ \emph{Add to buffer}}
\ENDFOR
\STATE {\color{CadetBlue} \emph{// Update TOLD using collected data in $\mathcal{B}$:}}
\FOR{num updates per episode}
\STATE $\{ \mathbf{s}_{t}, \mathbf{a}_{t}, r_{t}, \mathbf{s}_{t+1} \}_{t:t+H} \sim \mathcal{B}$~~~~~~~~~~~~~{\color{CadetBlue}$\vartriangleleft$ \emph{Sample traj.}}
\STATE $\mathbf{z}_{t} = h_{\theta}(\mathbf{s}_{t})$~~~~~~~~~~~~~~~~~~~~~{\color{CadetBlue}$\vartriangleleft$ \emph{Encode first observation}}
\STATE $J = 0$~~~~~~~~~~~~~~~~~~{\color{CadetBlue}$\vartriangleleft$ \emph{Initialize $J$ for loss accumulation}}
\FOR{$i = t...t+H$}
\STATE $\hat{r}_{i} = R_{\theta}(\mathbf{z}_{i}, \mathbf{a}_{i})$~~~~~~~~~~~~~~~~~~~~~~~~~~~~~{\color{CadetBlue}$\vartriangleleft$ \emph{Equation \ref{eq:reward-loss}}}
\STATE $\hat{q}_{i} = Q_{\theta}(\mathbf{z}_{i}, \mathbf{a}_{i})$~~~~~~~~~~~~~~~~~~~~~~~~~~~~{\color{CadetBlue}$\vartriangleleft$ \emph{Equation \ref{eq:value-loss}}}
\STATE $\mathbf{z}_{i+1} = d_{\theta}(\mathbf{z}_{i}, \mathbf{a}_{i})$~~~~~~~~~~~~~~~~~~~~~~~{\color{CadetBlue}$\vartriangleleft$ \emph{Equation \ref{eq:consistency-loss}}}
\STATE $\hat{a}_{i} = \pi_{\theta}(\mathbf{z}_{i})$~~~~~~~~~~~~~~~~~~~~~~~~~~~~~~~~~~{\color{CadetBlue}$\vartriangleleft$ \emph{Equation \ref{eq:policy-objective}}}
\STATE $J \leftarrow J + \lambda^{i-t} \mathcal{L}(\mathbf{z}_{i+1}, \hat{r}_{i}, \hat{q}_{i}, \hat{\mathbf{a}}_{i})$~~{\color{CadetBlue}$\vartriangleleft$ \emph{Equation \ref{eq:objective}}}
\ENDFOR
\STATE $\theta \leftarrow \theta - \frac{1}{H} \eta \nabla_{\theta} J$~~~~~~~~~~~~~~{\color{CadetBlue}$\vartriangleleft$ \emph{Update online network}}
\STATE $\theta^{-} \leftarrow (1 - \tau) \theta^{-} + \tau \theta$~~~~~{\color{CadetBlue}$\vartriangleleft$ \emph{Update target network}}
\ENDFOR
\ENDWHILE
\end{algorithmic}
\end{algorithm}
\vspace{-0.35in}
\end{figure}

\begin{figure*}[t]
    \centering
    \includegraphics[width=0.88\textwidth]{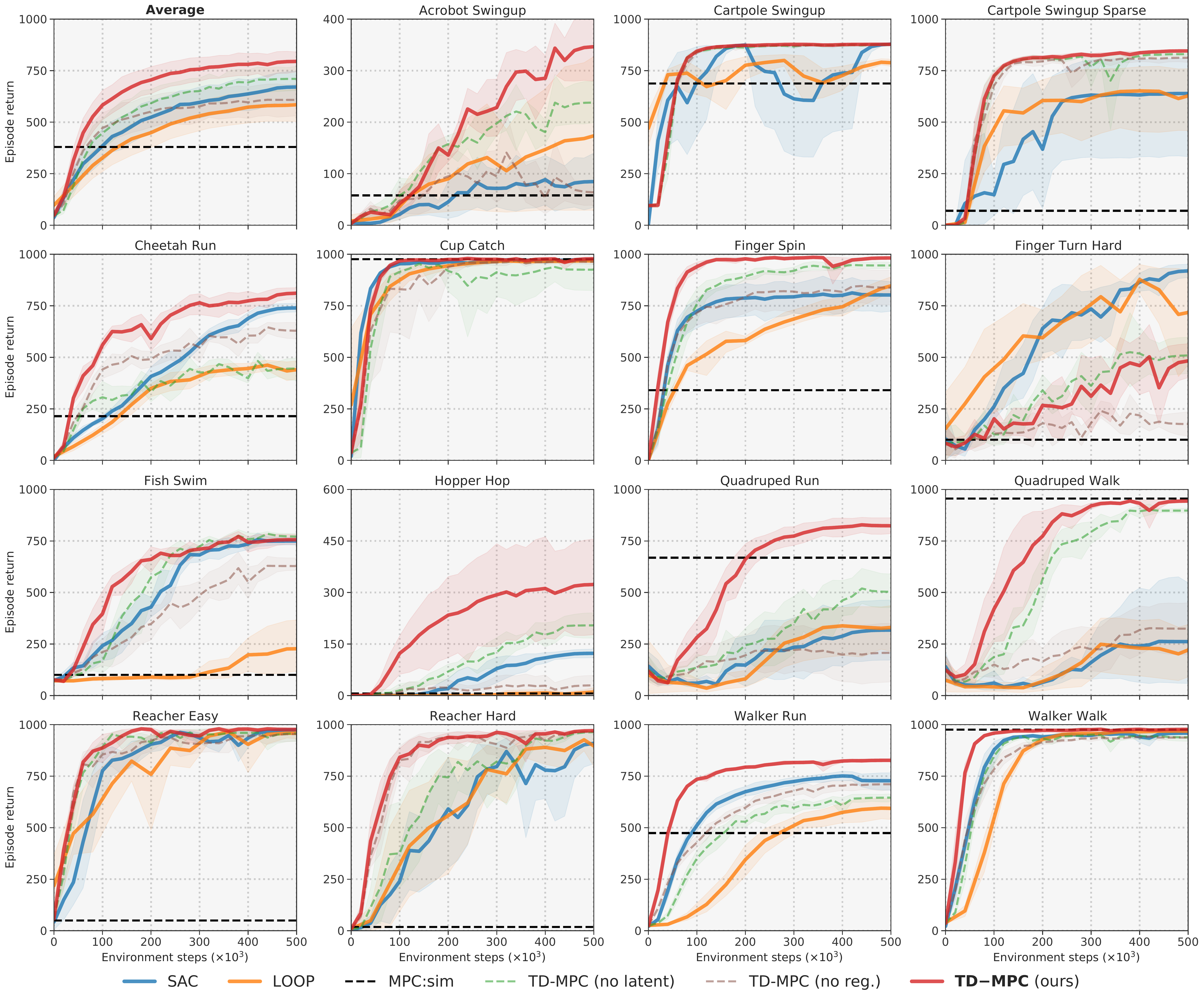}
    \vspace{-0.2in}
    \caption{\textbf{DMControl tasks.} Return of our method (TD-MPC) and baselines on 15 state-based continuous control tasks from DMControl \citep{deepmindcontrolsuite2018}. Mean of 5 runs; shaded areas are $95\%$ confidence intervals. In the top left, we visualize results averaged across all 15 tasks. We observe especially large performance gains on tasks with complex dynamics, e.g., the \textit{Quadruped} and \textit{Acrobot} tasks.}
    \label{fig:dmcontrol-state}
    \vspace{-0.2in}
\end{figure*}

\vspace{-0.025in}
\section{Experiments}
\label{sec:experiments}
\vspace{-0.025in}
We evaluate TD-MPC with a TOLD model on a total of $\mathbf{92}$ diverse and challenging continuous control tasks from DeepMind Control Suite (DMControl; \citet{deepmindcontrolsuite2018}) and Meta-World \textit{v2} \citep{yu2019meta}, including tasks with sparse rewards, high-dimensional state and action spaces, image observations, multi-modal inputs, goal-conditioning, and multi-task learning settings; see Appendix \ref{sec:appendix-visualizations} for task visualizations. We choose these two benchmarks for their great task diversity and availability of baseline implementations and results. We seek to answer the following questions:

$\boldsymbol{-}$ How does planning with TD-MPC compare to state-of-the-art model-based and model-free approaches?\vspace{0.05in}\\
$\boldsymbol{-}$ Are TOLD models capable of multi-task and transfer behaviors despite using a reward-centric objective?\vspace{0.05in}\\
$\boldsymbol{-}$ How does performance relate to the computational budget of the planning procedure?

An implementation of TD-MPC is available at \url{https://nicklashansen.github.io/td-mpc}, which will solve most tasks in an hour on a single GPU.

\textbf{Implementation details.} All components are deterministic and implemented using MLPs. We linearly anneal the exploration parameter $\epsilon$ of $\Pi_{\theta}$ and $\pi_{\theta}$ from $0.5$ to $0.05$ over the first 25k decision steps\footnote{To avoid ambiguity, we refer to simulation steps as \textit{environment steps} (independent of action repeat), and use \textit{decision steps} when referring to policy queries (dependent on action repeat).}. We use a planning horizon of $H=5$, and sample trajectories using prioritized experience replay \citep{Schaul2016PrioritizedER} with priority scaled by the value loss. During planning, we plan for $6$ iterations ($8$ for Dog; $12$ for Humanoid), sampling $N=512$ trajectories ($+5\%$ sampled from $\pi_{\theta}$), and we compute $\mu,\sigma$ parameters over the top-$64$ trajectories each iteration. For image-based tasks, observations are 3 stacked $84\times 84$-dimensional RGB frames and we use $\pm4$ pixel shift augmentation \citep{kostrikov2020image}. Refer to Appendix \ref{sec:appendix-implementation-details} for additional details.

\textbf{Baselines.} We evaluate our method against the following:

$\boldsymbol{-}$ \textbf{Soft Actor-Critic} (SAC; \citep{Haarnoja2018SoftAA}), a state-of-the-art model-free algorithm derived from maximum entropy RL \citep{ziebart2008maximum}. We choose SAC as our main point of comparison due to its popularity and strong performance on both DMControl and Meta-World. In particular, we adopt the implementation of \citet{pytorch_sac}.\vspace{0.05in}\\
$\boldsymbol{-}$ \textbf{LOOP} \citep{Sikchi2020LearningOW}, a hybrid algorithm that extends SAC with planning and a learned model. LOOP has been shown to outperform a number of model-based methods, e.g., MBPO \citep{Janner2019WhenTT} and POLO \citep{Lowrey2019PlanOL}) on select MuJoCo tasks. It is a particularly relevant baseline due to its similarities to TD-MPC.\vspace{0.05in}\\
$\boldsymbol{-}$ \textbf{MPC} with a \textit{ground-truth} simulator (denoted \textit{MPC:sim}). As planning with a simulator is computationally intensive, we limit the planning horizon to $10$ ($2\times$ ours), sampled trajectories to 200, and optimize for 4 iterations (ours: 6).\vspace{0.05in}\\
$\boldsymbol{-}$ \textbf{CURL} \citep{srinivas2020curl}, \textbf{DrQ} \citep{kostrikov2020image}, and \textbf{DrQ-v2} \citep{yarats2021drqv2}, three state-of-the-art model-free algorithms.
\vspace{0.05in}\\
$\boldsymbol{-}$ \textbf{PlaNet} \citep{hafner2019planet}, \textbf{Dreamer} \citep{Hafner2020DreamTC}, and \textbf{Dreamer-v2} \citep{hafner2020dreamerv2}. All three methods learn a model using a reconstruction loss, and select actions using either MPC or a learned policy.\vspace{0.05in}\\
$\boldsymbol{-}$ \textbf{MuZero} \citep{Schrittwieser2020MasteringAG} and \textbf{EfficientZero} \citep{Ye2021MasteringAG}, which learn a latent dynamics model from rewards and uses MCTS for discrete action selection.\vspace{0.05in}\\
$\boldsymbol{-}$ \textbf{Ablations}. We consider: \textit{(i)} our method implemented using a state predictor ($h_{\theta}$ being the identity function), \textit{(ii)} our method implemented without the latent consistency loss from Equation \ref{eq:consistency-loss}, and lastly: the consistency loss replaced by either \textit{(iii)} the \emph{reconstruction} objective of PlaNet and Dreamer, or \textit{(iv)} the \emph{contrastive} objective of EfficientZero.

\input{tables/planet}

See Appendix \ref{sec:appendix-baselines} for further discussion on baselines.

\begin{figure}[t]
    \centering
    \includegraphics[width=0.3985\textwidth]{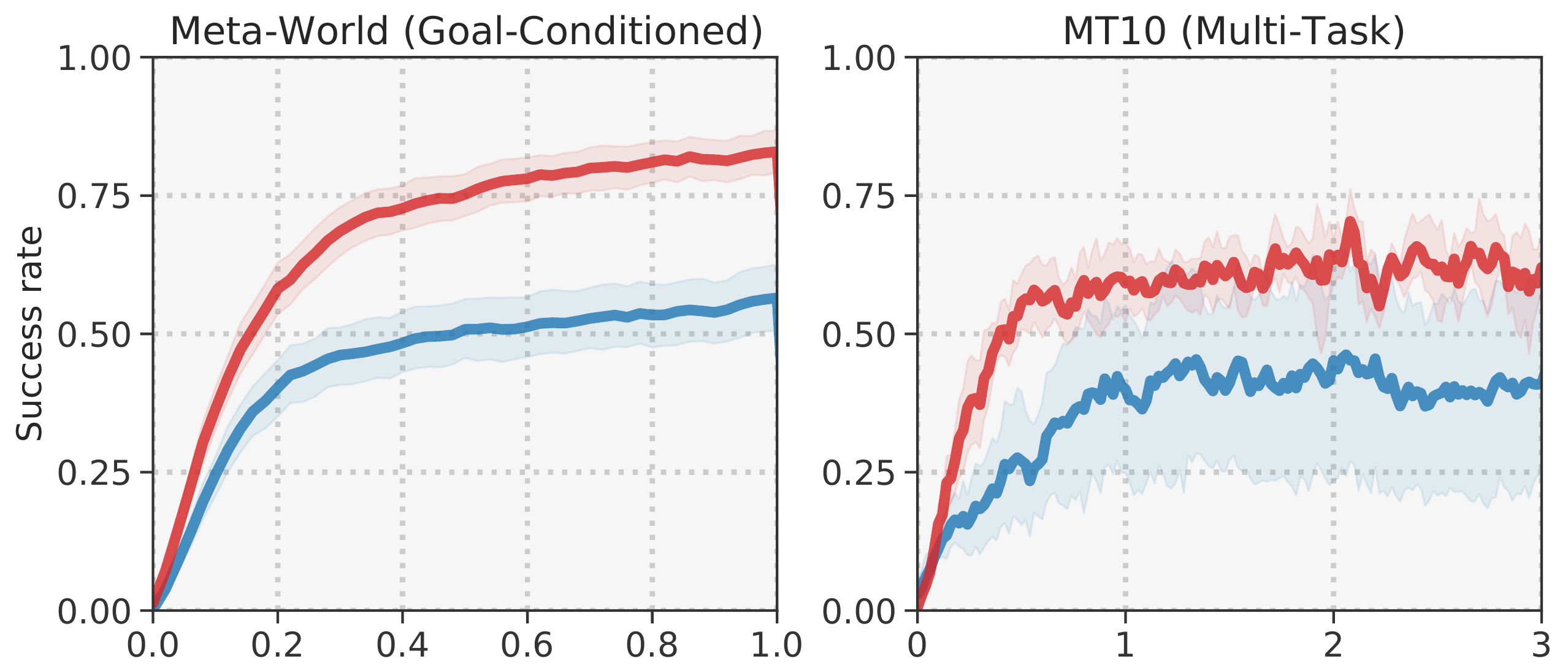}\vspace{0.025in}
    \includegraphics[width=0.485\textwidth]{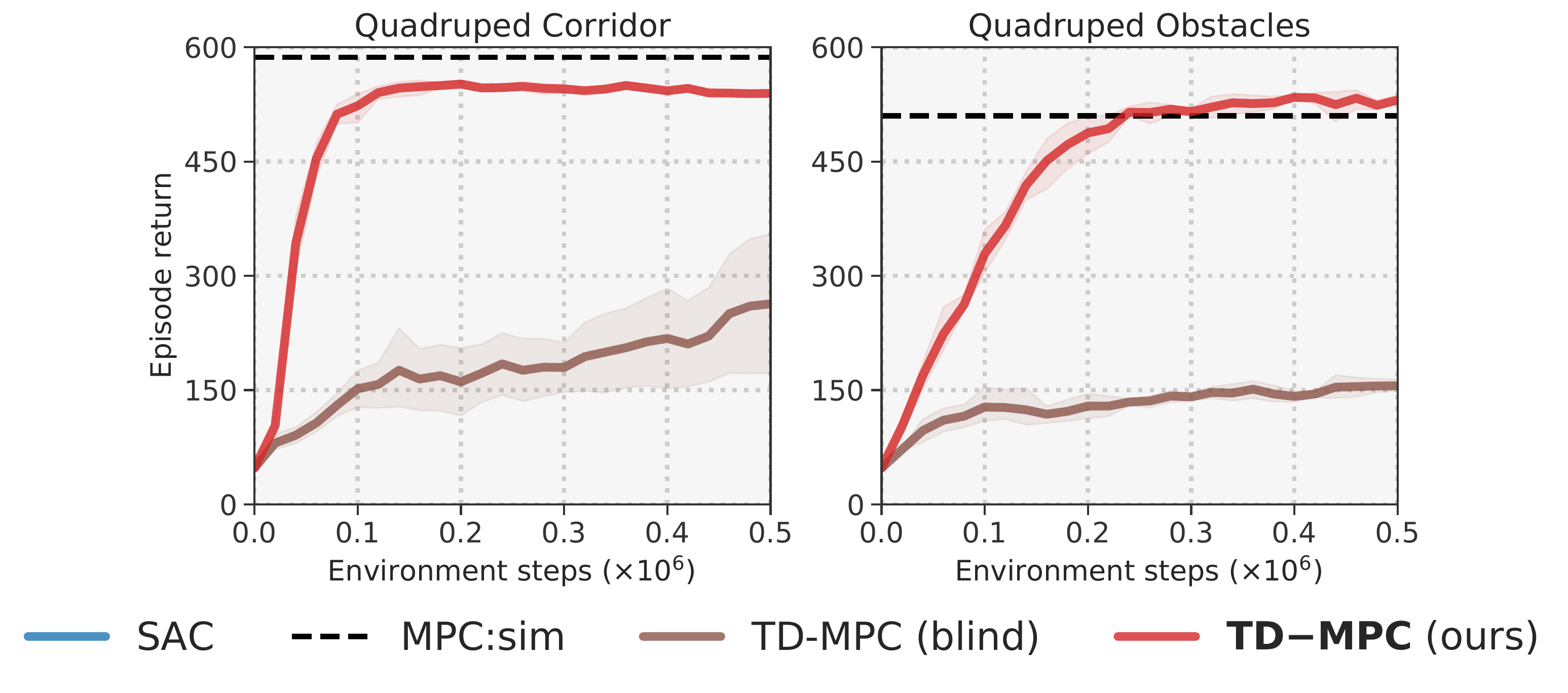}
    \vspace{-0.35in}
    \caption{\textit{(top)} \textbf{Meta-World.} Success rate on \textbf{50} goal-conditioned Meta-World tasks using individual policies, and a multi-task policy trained on 10 tasks simultaneously (Meta-World MT10). Individual task results shown in Appendix \ref{sec:appendix-metaworld}. \textit{(bottom)} \textbf{Multi-modal RL.} Episode return of TD-MPC on two multi-modal locomotion tasks using proprioceptive data + an egocentric camera. \textit{Blind} uses only proprioceptive data. See Appendix \ref{sec:appendix-visualizations} for visualizations. All results are means of $5$ runs; shaded areas are $95\%$ confidence intervals.}
    \label{fig:metaworld-all}
    \vspace{-0.1in}
\end{figure}

\textbf{Tasks.} We consider the following \textbf{92} tasks:

$\boldsymbol{-}$ $\mathbf{6}$ challenging Humanoid ($\mathcal{A} \in \mathbb{R}^{21}$) and Dog ($\mathcal{A} \in \mathbb{R}^{38}$) locomotion tasks with high-dimensional state and action spaces. Results are shown in Figure \ref{fig:planning}.\vspace{0.05in}\\
$\boldsymbol{-}$ $\mathbf{15}$ diverse continuous control tasks from DMControl, 6 of which have sparse rewards. Results shown in Figure \ref{fig:dmcontrol-state}.\vspace{0.05in}\\
$\boldsymbol{-}$ $\mathbf{6}$ image-based tasks from the data-efficient DMControl 100k benchmark. Results are shown in Table \ref{tab:planet}.\vspace{0.05in}\\
$\boldsymbol{-}$ $\mathbf{12}$ image-based tasks from the DMControl \textit{Dreamer} benchmark (3M environment steps). Results in Figure \ref{fig:dreamer}.\vspace{0.05in}\\
$\boldsymbol{-}$ $\mathbf{2}$ multi-modal (proprioceptive data + egocentric camera) 3D locomotion tasks in which a quadruped agent navigates around obstacles. Results are shown in Figure \ref{fig:metaworld-all} (middle).\vspace{0.05in}\\
$\boldsymbol{-}$ $\mathbf{50}$ goal-conditioned manipulation tasks from Meta-World, as well as a multi-task setting where 10 tasks are learned simultaneously. Results are shown in Figure \ref{fig:metaworld-all} (top).

Throughout, we benchmark performance on relatively few environment steps, e.g., 3M steps for Humanoid tasks whereas prior work typically runs for 30M steps ($10\times$).

\textbf{Comparison to other methods.} We find our method to outperform or match baselines in most tasks considered, generally with larger gains on complex tasks such as \emph{Humanoid}, \emph{Dog} (DMControl), and \emph{Bin Picking} (Meta-World), and we note that TD-MPC is in fact \emph{\textbf{the first documented result}} solving the complex \emph{Dog} tasks of DMControl. Performance of LOOP is similar to SAC, and MPC with a simulator (\textit{MPC:sim}) performs well on locomotion tasks but fails in tasks with sparse rewards. Although we did not tune our method specifically for image-based RL, we obtain results competitive with state-of-the-art model-based and model-free algorithms that are both carefully tuned for image-based RL and contain up to $15\times$ more learnable parameters. Notably, while EfficientZero produces strong results on tasks with low-dimensional action spaces, its Monte-Carlo Tree Search (MCTS) requires discretization of action spaces, which is unfeasible in high dimensions. In contrast, TD-MPC scales remarkably well to the 38-dimensional continuous action space of \emph{Dog} tasks. Lastly, we observe inferior sample efficiency compared to SAC and LOOP on the hard exploration task \textit{Finger Turn Hard} in Figure \ref{fig:dmcontrol-state}, which suggests that incorporating more sophisticated exploration strategies might be promising for future research. We defer experiments that ablate the choice of regularization loss to Appendix \ref{sec:appendix-loss-ablations}, but find our proposed latent state consistency loss to yield the most consistent results.

\textbf{Multi-task RL, multi-modal RL, and generalization.} A common argument in favor of general-purpose models is that they can benefit from data-sharing across tasks. Therefore, we seek to answer the following question: does TOLD similarly benefit from synergies between tasks, despite its reward-centric objective? We test this hypothesis through two experiments: training a single policy to perform 10 different tasks simultaneously (Meta-World MT10), and evaluating model generalization when trained on one task (\textit{Walk}) and transferring to a different task from the same domain (\textit{Run}). Multi-task results are shown in Figure \ref{fig:metaworld-all} (top), and transfer results are deferred to Appendix \ref{sec:appendix-transfer}. We find our method to benefit from data sharing in both experiments, and our transfer results indicate that $h_{\theta}$ generalizes well to new tasks, while $d_{\theta}$ encodes more task-specific behavior. We conjecture that, while TOLD only learns features that are predictive of reward, similar tasks often have similar reward structures, which enables sharing of information between tasks. However, we still expect general-purpose models to benefit more from \textit{unrelated} tasks in the same environment than TOLD. Lastly, an added benefit of our task-centric objective is that it is agnostic to the input modality. To demonstrate this, we solve two multi-modal (proprioceptive data + egocentric camera) locomotion tasks using TD-MPC; results in Figure \ref{fig:metaworld-all} (bottom). We find that TD-MPC successfully fuses information from the two input modalities, and solves the tasks. In contrast, a blind agent that does \textit{not} have access to the egocentric camera fails. See Appendix \ref{sec:appendix-multimodal} for further details on the multi-modal experiments, and Appendix \ref{sec:appendix-metaworld} for details on the multi-task experiments.

\begin{figure}[t]
    \centering
    \includegraphics[width=0.48\textwidth]{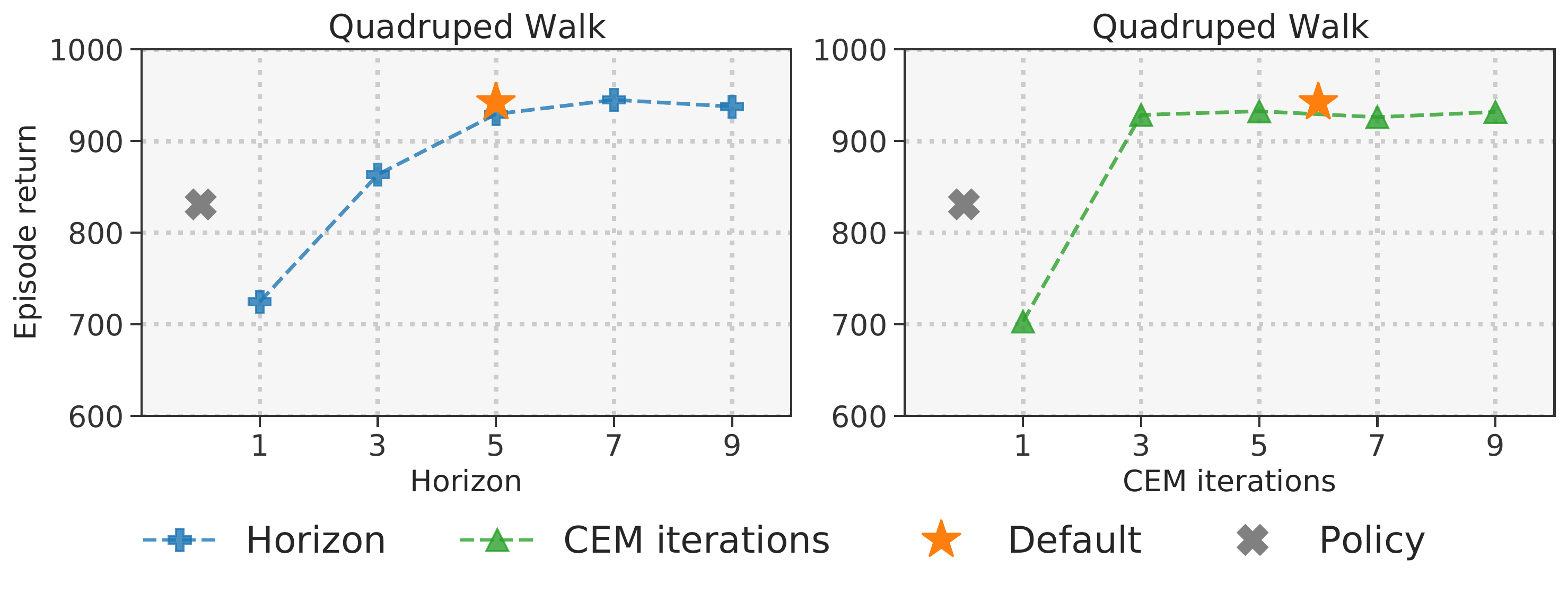}
    \vspace{-0.375in}
    \caption{\textbf{Variable computational budget.} Return of TD-MPC on \textit{Quadruped Walk} under a variable budget. We evaluate performance of fully trained agents when varying \textit{(left)} planning horizon; \textit{(right)} number of iterations during planning. When varying one hyperparameter, the other is fixed to the default value. We include evaluation of the learned policy $\pi_{\theta}$, and the default setting of $6$ iterations and a horizon of $5$ used in training. Mean of 5 runs.
    }
    \label{fig:computational-budget}
    \vspace{-0.125in}
\end{figure}

\textbf{Performance vs. computational budget.} We investigate the relationship between computational budget (i.e., planning horizon and number of iterations) and performance in DMControl tasks; see Figure \ref{fig:computational-budget}. We find that, for complex tasks such as \textit{Quadruped Walk} ($\mathcal{A} \in \mathbb{R}^{12}$), more planning generally leads to better performance. However, we also observe that we can reduce the planning cost during inference by $\mathbf{50\%}$ (compared to during training) \textit{without} a drop in performance by reducing the number of iterations. For particularly fast inference, one can discard planning altogether and simply use the jointly learned policy $\pi_{\theta}$; however, $\pi_{\theta}$ generally performs worse than planning. See Appendix \ref{sec:appendix-variable-compute} for additional results.

\textbf{Training wall-time.} To better ground our results, we report the training wall-time of TD-MPC compared to SAC, LOOP that is most similar to our method, and MPC with a ground-truth simulator (non-parametric). Methods are benchmarked on a single RTX3090 GPU. Results are shown in Table \ref{tab:walltime}. TD-MPC solves \textit{Walker Walk} $\mathbf{16}\times$ faster than LOOP and matches the time-to-solve of SAC on both \textit{Walker Walk} and \textit{Humanoid Stand} while being significantly more sample efficient. Thus, our method effectively closes the time-to-solve gap between model-free and model-based methods. This is a nontrivial reduction, as LOOP is already known to be, e.g., $12\times$ faster than the purely model-based method, POLO \citep{Lowrey2019PlanOL, Sikchi2020LearningOW}. We provide additional experiments on inference times in Appendix \ref{sec:appendix-inference}.

\input{tables/walltime}

\vspace{-0.025in}
\section{Related Work}
\label{sec:related-work}
\vspace{-0.05in}
\textbf{Temporal Difference Learning.} Popular model-free off-policy algorithms such as DDPG \citep{Lillicrap2016ContinuousCW} and SAC \citep{Haarnoja2018SoftAA} represent advances in deep TD-learning based on a large body of literature \citep{Sutton1988TD, mnih2013playing, Hasselt2016DeepRL, Mnih2016AsynchronousMF, Fujimoto2018AddressingFA, Kalashnikov2018QTOptSD, Espeholt2018IMPALASD, Pourchot2019CEMRLCE, Kalashnikov2021MTOptCM}. Both DDPG and SAC learn a policy $\pi_{\theta}$ and value function $Q_{\theta}$, but do not learn a model. \citet{Kalashnikov2018QTOptSD, shao2020grac, Kalashnikov2021MTOptCM} also learn $Q_{\theta}$, but replace or augment $\pi_{\theta}$ with model-free CEM. Instead, we jointly learn a model, value function, and policy using TD-learning, and interact using sampling-based planning.

\textbf{Model-based RL.} A common paradigm is to learn a model of the environment that can be used for planning \citep{Ebert2018VisualFM, Zhang2018SOLARDS, Janner2019WhenTT, hafner2019planet, Lowrey2019PlanOL, Kaiser2020ModelBasedRL, Bhardwaj2020InformationTM, Yu2020MOPOMO, Schrittwieser2020MasteringAG, Nguyen2021TemporalPC} or for training a model-free algorithm with generated data \citep{Pong2018TemporalDM, ha2018worldmodels, Hafner2020DreamTC, Sekar2020PlanningTE}. For example, \citet{Zhang2018SOLARDS, ha2018worldmodels, hafner2019planet, Hafner2020DreamTC} learn a dynamics model using a video prediction loss, \citet{Yu2020MOPOMO, Kidambi2020MOReLM} consider model-based RL in the offline setting, and MuZero/EfficientZero \citep{Schrittwieser2020MasteringAG,Ye2021MasteringAG} learn a latent dynamics model using reward prediction. EfficientZero is most similar to ours in terms of model learning, but its MCTS-based action selection is inherently incompatible with continuous action spaces. Finally, while learning a terminal value function for MPC has previously been proposed \citep{negenborn2005, Lowrey2019PlanOL, Bhardwaj2020InformationTM, Hatch2021TheVO}, we are (to the best of our knowledge) the first to jointly learn model and value function through TD-learning in continuous control.

\textbf{Hybrid algorithms.} Several prior works aim to develop algorithms that combine model-free and model-based elements \citep{Nagabandi2018NeuralND, Buckman2018SampleEfficientRL, Pong2018TemporalDM, Hafez2019CuriousMA, Sikchi2020LearningOW, Wang2020ExploringMP, Clavera2020ModelAugmentedAB, hansen2021deployment, Morgan2021ModelPA, Bhardwaj2021BlendingM, Margolis2021LearningTJ}, many of which are orthogonal to our contributions. For example, \citet{Clavera2020ModelAugmentedAB} and \citet{Buckman2018SampleEfficientRL, Lowrey2019PlanOL} use a learned model to improve policy and value learning, respectively, through generated trajectories. LOOP \citep{Sikchi2020LearningOW} extends SAC with a learned state prediction model and constrains planned trajectories to be close to those of SAC, whereas we \emph{replace} the parameterized policy by planning with TD-MPC and learn a \emph{task-oriented} latent dynamics model.

We provide a qualitative comparison of key components in TD-MPC and prior work in Appendix \ref{sec:comparison-related-work}.

\section{Conclusions and Future Directions}
\label{sec:conclusion}
We are excited that our TD-MPC framework, despite being markedly distinct from previous work in the way that the model is learned and used, is already able to outperform model-based and model-free methods on diverse continuous control tasks, and (with trivial modifications) simultaneously match state-of-the-art on image-based RL tasks. Yet, we believe that there is ample opportunity for performance improvements by extending the TD-MPC framework. For example, by using the learned model in creative ways \citep{Clavera2020ModelAugmentedAB, Buckman2018SampleEfficientRL, Lowrey2019PlanOL}, incorporating better exploration strategies, or improving the model through architectural innovations.

\subsection*{Acknowledgements}
\label{sec:acknowledgements}
\vspace{-0.025in}
This project is supported, in part, by grants from NSF CCF-2112665 (TILOS), and gifts from Meta, Qualcomm.

The authors would like to thank Yueh-Hua Wu, Ruihan Yang, Sander Tonkens, Tongzhou Mu, and Yuzhe Qin for helpful discussions.

\newpage

\bibliographystyle{icml2022}
\bibliography{main}

\appendix

\input{tables/related-work}

\begin{figure}[h]
    \centering
    \includegraphics[width=0.44\textwidth]{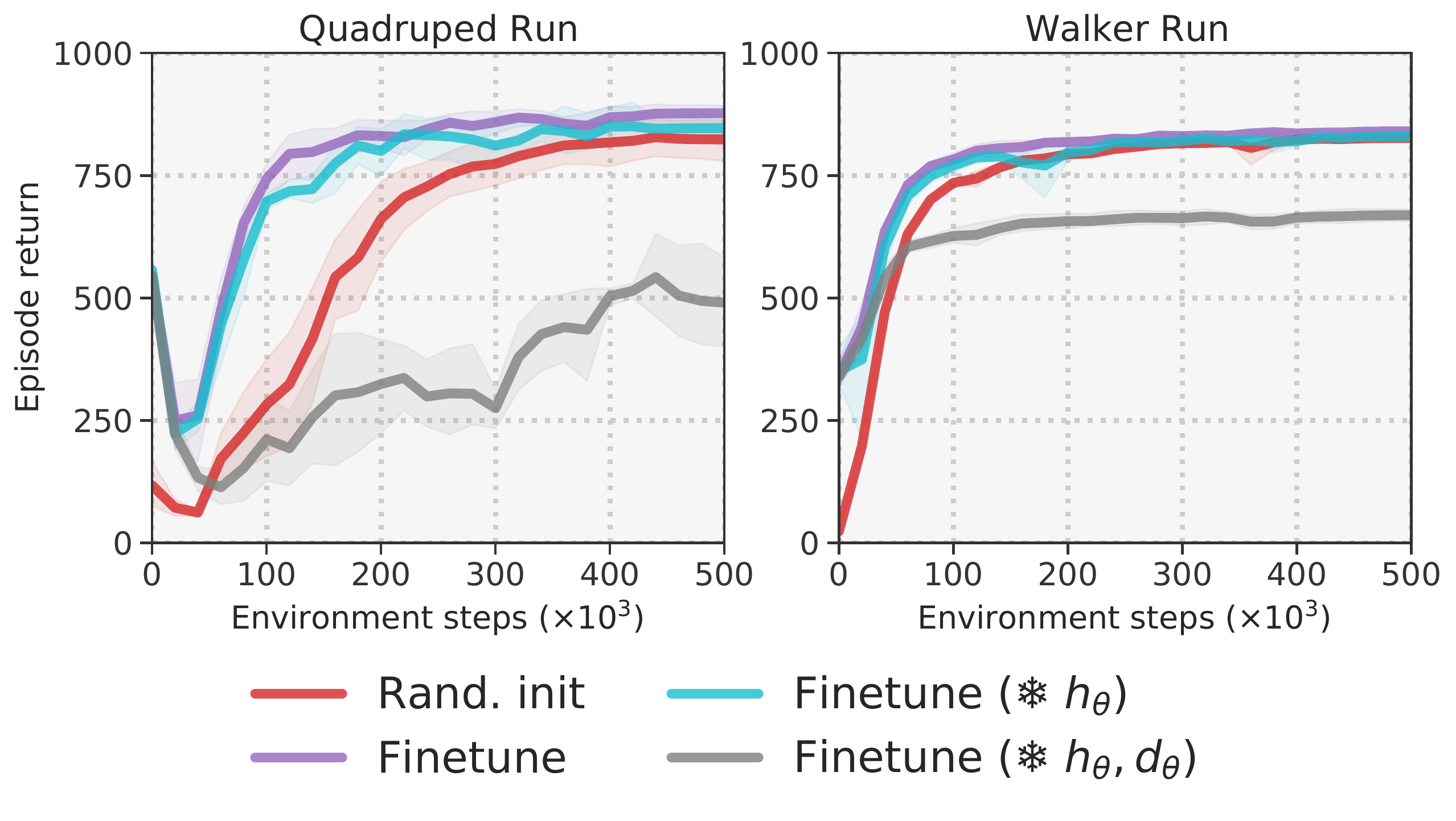}
    \vspace{-0.2in}
    \caption{\textbf{Model generalization.} Return of our method under three different settings: \textit{(Rand. init)} TD-MPC trained from scratch on the two \emph{Run} tasks; \textit{(Finetune)} TD-MPC initially trained on \emph{Walk} tasks and then finetuned online on \emph{Run} tasks without any weights frozen;  \textit{(Finetune, freeze $h_{\theta}$)} same setting as before, but with the encoder $h_{\theta}$ frozen; and \textit{(Finetune, freeze $h_{\theta}, d_{\theta}$)} both encoder $h_{\theta}$ and latent dynamics predictor $d_{\theta}$ frozen. Mean of 5 runs; shaded areas are $95\%$ confidence intervals.}
    \label{fig:appendix-transfer}
    \vspace{-0.1in}
\end{figure}

\begin{figure*}[t]
    \centering
    \includegraphics[width=0.48\textwidth]{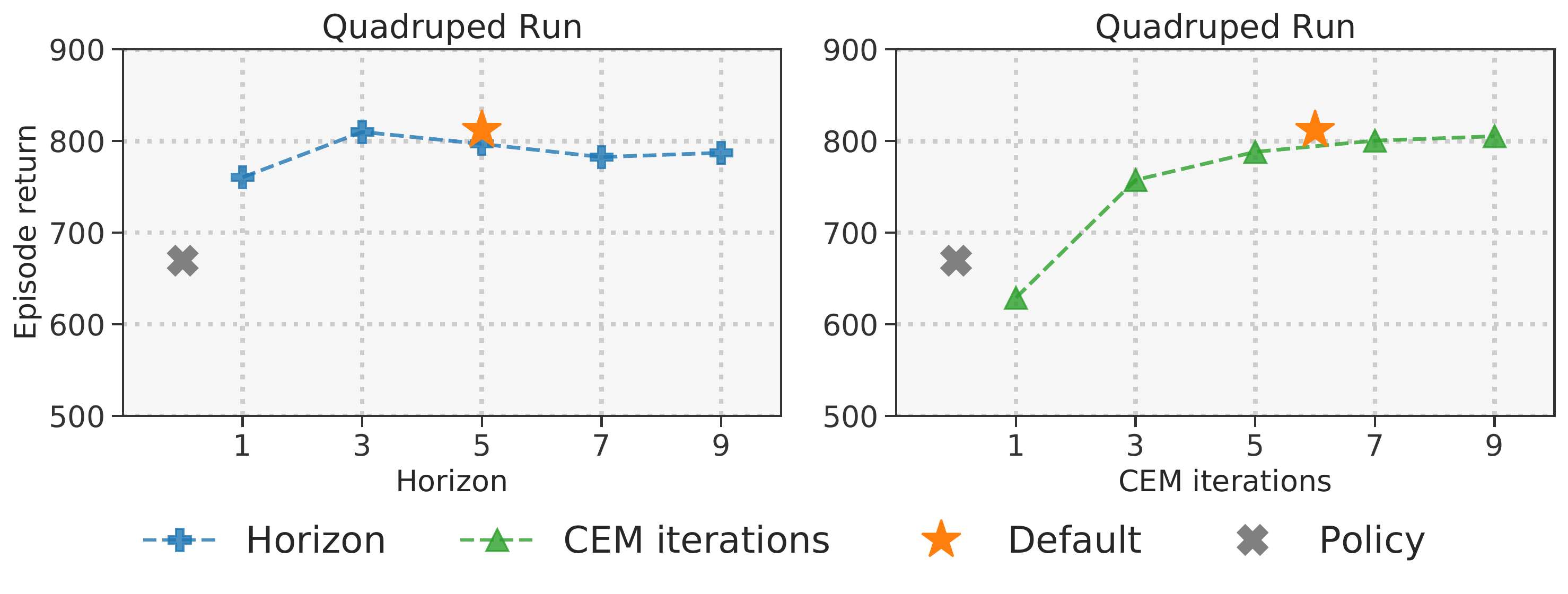}\quad
    \includegraphics[width=0.48\textwidth]{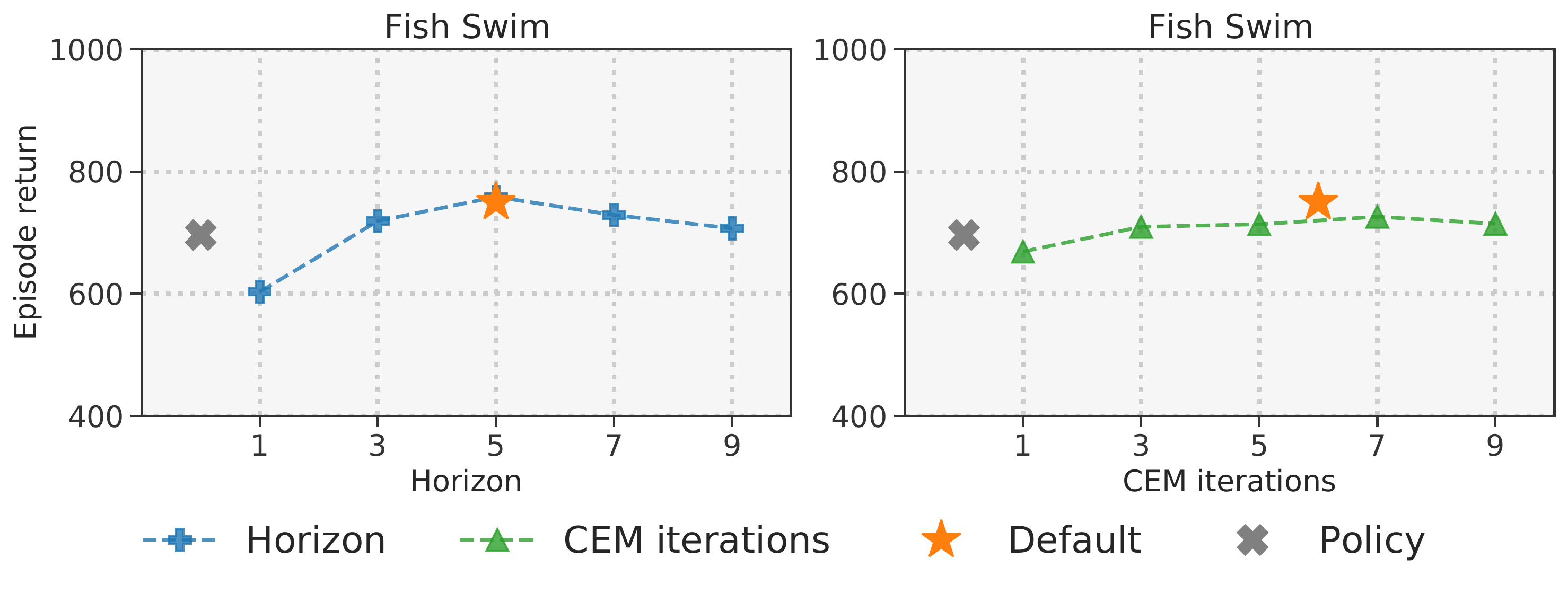}\\
    \includegraphics[width=0.48\textwidth]{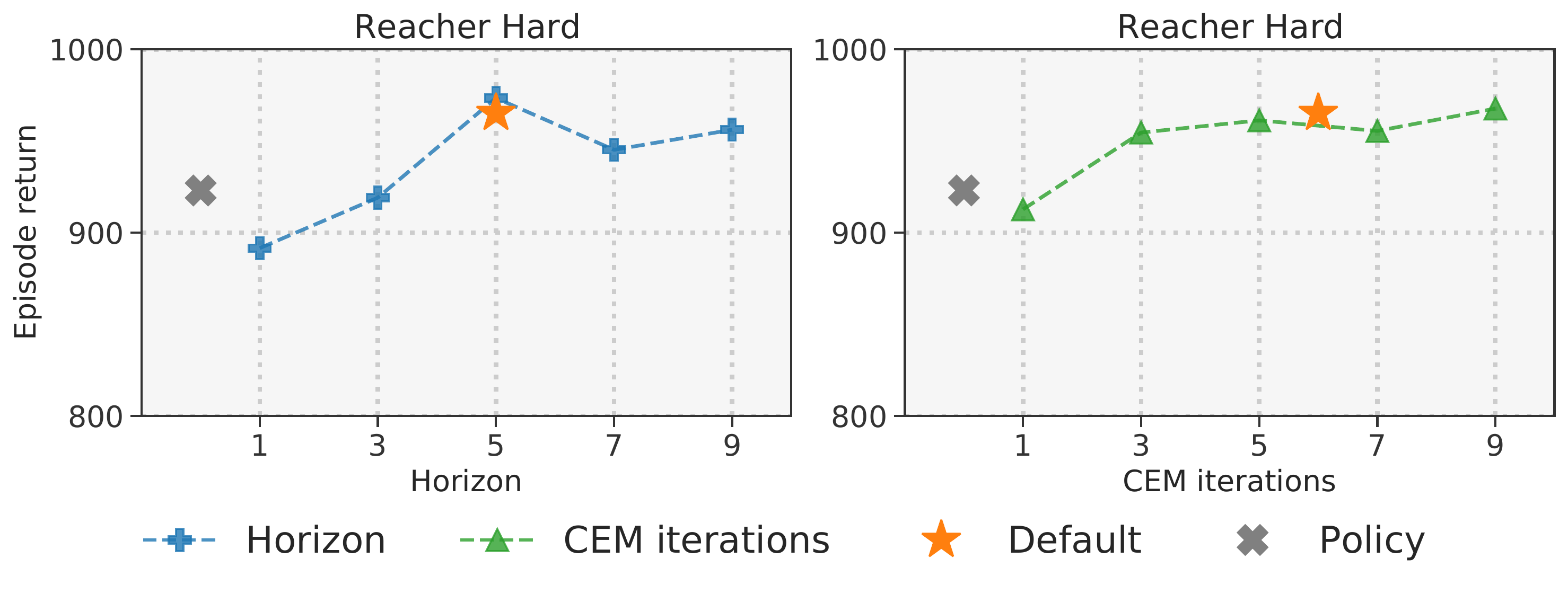}\quad
    \includegraphics[width=0.48\textwidth]{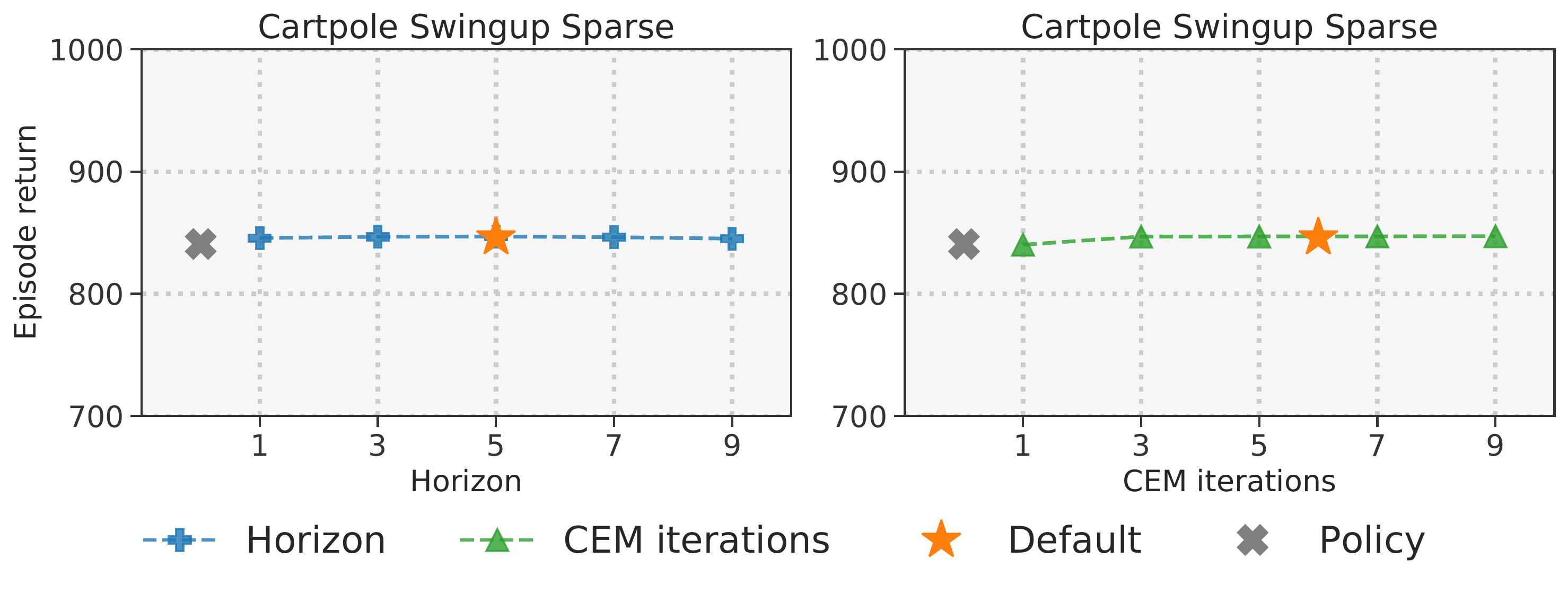}
    \vspace{-0.15in}
    \caption{\textbf{Variable computational budget.} Return of our method (TD-MPC) under a variable computational budget. In addition to the task in Figure \ref{fig:computational-budget}, we provide results on four other tasks from DMControl: \textit{Quadruped Run} ($\mathcal{A} \in \mathbb{R}^{12}$), \textit{Fish Swim} ($\mathcal{A} \in \mathbb{R}^{5}$), \textit{Reacher Hard} ($\mathcal{A} \in \mathbb{R}^{2}$), and \textit{Cartpole Swingup Sparse} ($\mathcal{A} \in \mathbb{R}$). We evaluate performance of fully trained agents when varying \textit{(blue)} planning horizon; \textit{(green)} number of iterations during planning. For completeness, we also include evaluation of the jointly learned policy $\pi_{\theta}$, as well as the default setting of $6$ iterations and a horizon of $5$ used during training. Higher values require more compute. Mean of 5 runs. }
    \label{fig:appendix-computational-budget}
    \vspace{-0.05in}
\end{figure*}

\begin{figure}[t]
    \centering
    \includegraphics[width=0.34\textwidth]{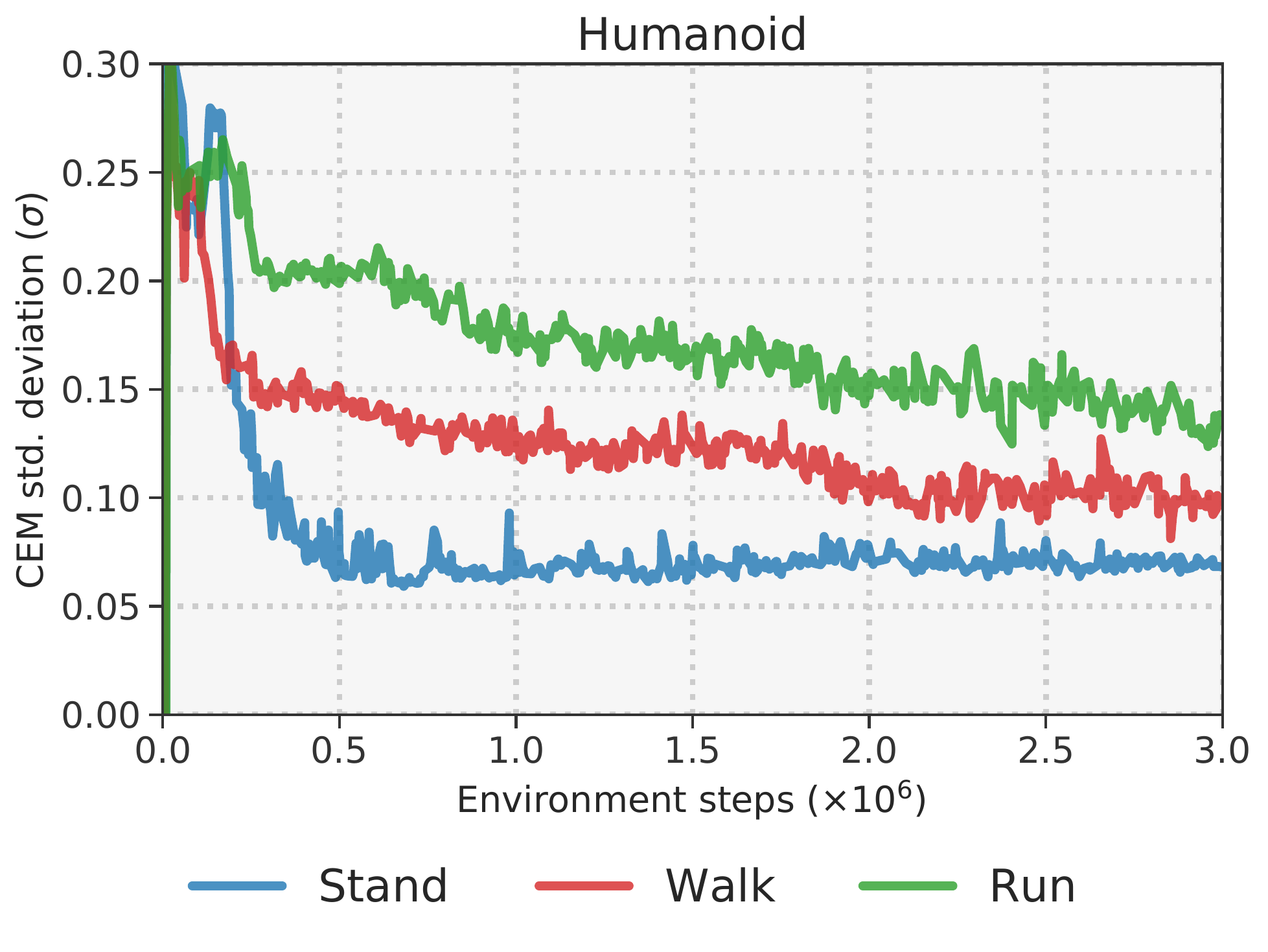}
    \vspace{-0.2in}
    \caption{\textbf{Exploration by planning.} Average std. deviation ($\sigma$) of our planning procedure after the final iteration of planning over the course of training. Results are shown for the three Humanoid tasks: Stand, Walk, and Run, listed in order of increasing difficulty.}
    \label{fig:appendix-exploration}
    \vspace{-0.15in}
\end{figure}

\begin{figure*}[t]
    \centering
    \includegraphics[width=0.83\textwidth]{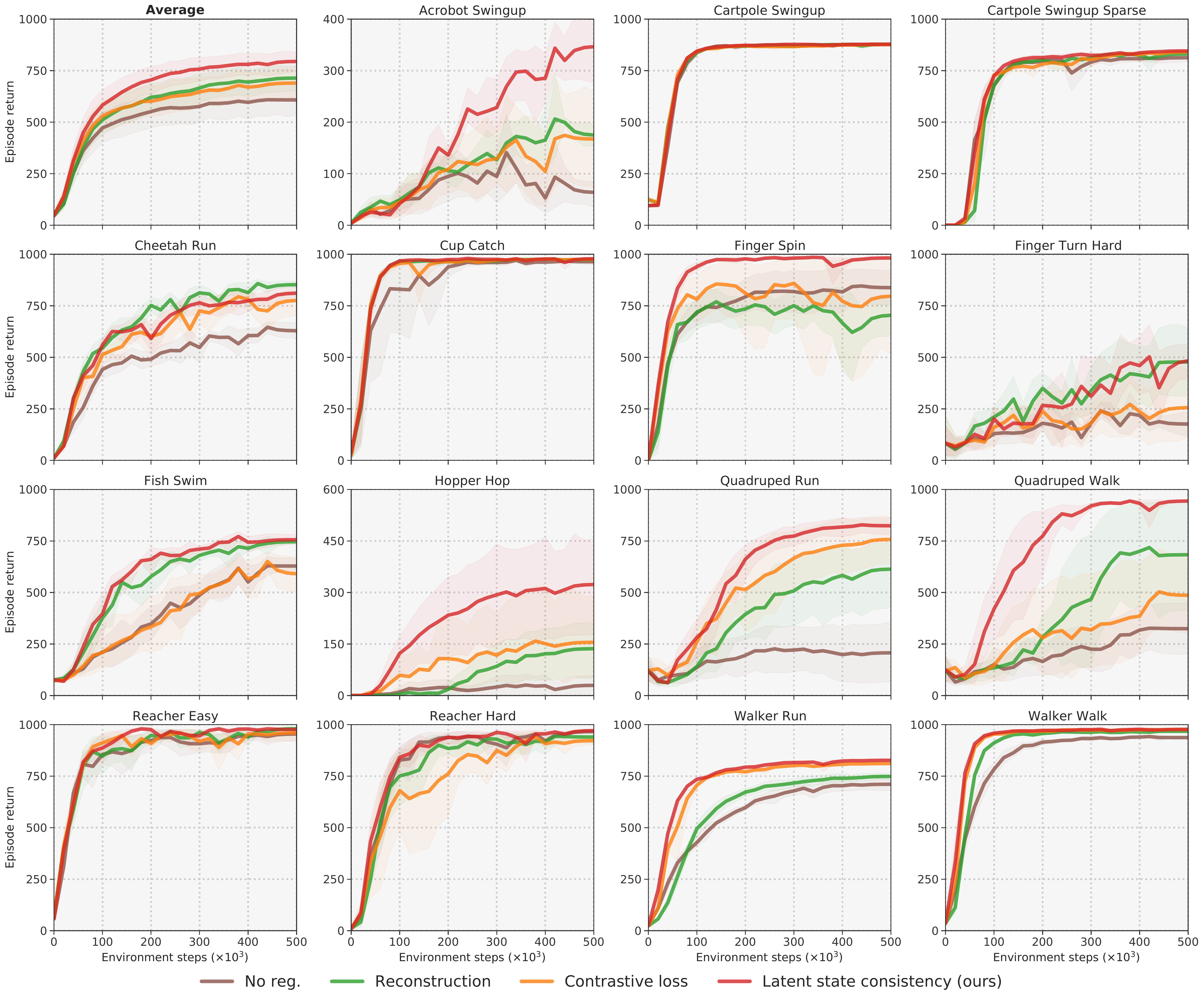}
    \vspace{-0.2in}
    \caption{\textbf{Latent dynamics objective.} Return of our method (TD-MPC) using different latent dynamics objectives in addition to reward and value prediction. 15 state-based continuous control tasks from DMControl \citep{deepmindcontrolsuite2018}. \textit{No reg.} uses no regularization term, \textit{reconstruction} uses a state prediction loss, \textit{contrastive loss} adopts the contrastive objective of \citet{Ye2021MasteringAG, hansen2021softda}, and \textit{latent state consistency} corresponds to Equation \ref{eq:consistency-loss}. Mean of 5 runs; shaded areas are $95\%$ confidence intervals. In the top left, we visualize results averaged across all 15 tasks. Both reconstruction and contrastive losses improve over the baseline without regularization, but our proposed latent state consistency loss yields more consistent results.}
    \label{fig:dmcontrol-state-contrastive}
    \vspace{-0.1in}
\end{figure*}

\section{Model Generalization}
\label{sec:appendix-transfer}

We investigate the transferability of a TOLD model between related tasks. Specifically, we consider model transfer in two locomotion domains, \textit{Walker} and \textit{Quadruped}, where we first train policies on \textit{Walk} tasks and then finetune the learned model on \textit{Run} tasks. We finetune in an online setting, i.e., the only difference between training from scratch and finetuning is the weight initialization, and we keep all hyperparameters identical. Results from the experiment are shown in Figure \ref{fig:appendix-transfer}. When finetuning the full TOLD model, we find our method to converge considerably faster, suggesting that TOLD does indeed learn features that transfer between related tasks. We perform two additional finetuning experiments: freezing parameters of the representation $h_{\theta}$, and freezing parameters of both $h_{\theta}$ \textit{and} the latent dynamics predictor $d_{\theta}$ during finetuning. We find that freezing $h_{\theta}$ nearly matches our results for finetuning without frozen weights, indicating that $h_{\theta}$ learns to encode information that transfers between tasks. However, when finetuning with both $h_{\theta},d_{\theta}$ frozen, rate of convergence degrades substantially, which suggests that $d_{\theta}$ tends to encode more task-specific behavior.

\section{Comparison to Prior Work}
\label{sec:comparison-related-work}
We here extend our discussion of related work in Section \ref{sec:related-work}. Table \ref{tab:related-work} provides a qualitative comparison of key components of TD-MPC and prior model-based and model-free approaches, e.g., comparing model objectives, use of a (terminal) value function, and inference-time behavior. While different aspects of TD-MPC have been explored in prior work, we are the first to propose a complete framework for MPC with a model learned by TD-learning.

\section{Variable Computational Budget}
\label{sec:appendix-variable-compute}

This section supplements our experiments in Figure \ref{fig:computational-budget} on a variable computational budget for planning during inference; additional results are shown in Figure \ref{fig:appendix-computational-budget}. We observe that the gap between planning performance and policy performance tends to be larger for tasks with high-dimensional action spaces such as the two \textit{Quadruped} tasks. We similarly find that performance varies relatively little when the computational budget is changed for tasks with simple dynamics (e.g., \textit{Cartpole} tasks) compared to tasks with more complex dynamics. We find that our default hyperparameters ($H=5$ and 6 iterations; shown as a star in Figure \ref{fig:appendix-computational-budget}) strikes a good balance between compute and performance.

\section{Latent Dynamics Objective}
\label{sec:appendix-loss-ablations}
We ablate the choice of latent dynamics objective by replacing our proposed latent state consistency loss in Equation \ref{eq:consistency-loss} with \textit{(i)} a contrastive loss similar to that of \citet{Ye2021MasteringAG, hansen2021softda}, and \textit{(ii)} a reconstruction objective similar to that of \citet{ha2018worldmodels, hafner2019planet, Hafner2020DreamTC}. Specifically, for (i) we adopt the recently proposed SimSiam \citep{Chen2021ExploringSS} self-supervised framework and implement the projection layer as an MLP with 2 hidden layers and output size 32, and the predictor head is an MLP with 1 hidden layer. All layers use ELU activations and a hidden size of 256. Consistent with the public implementations of \citet{Ye2021MasteringAG, hansen2021softda}, we find it beneficial to apply BatchNorm in the projection and predictor modules. We also find that using a higher loss coefficient of $c_{3}=100$ (up from 2) produces slightly better results. For (ii) we implement the decoder for state reconstruction by mirroring the encoder; an MLP with 1 hidden layer and ELU activations. We also include a \textit{no regularization} baseline for completeness. Results are shown in Figure \ref{fig:dmcontrol-state-contrastive}.

\section{Exploration by planning}
\label{sec:appendix-exploration}

We investigate the role that planning by TD-MPC has in exploration. Figure \ref{fig:appendix-exploration} shows the average std. deviation of our planning procedure after the final iteration of planning for the three Humanoid tasks: Stand, Walk, and Run, listed in order of increasing difficulty. We observe that the std. deviation (and thus degree of exploration) is decreasing as training progresses, and converges as the task becomes solved. Generally, we find that exploration decreases slower for hard tasks, which we conjecture is due to larger variance in reward and value estimates. As such, the TD-MPC framework inherently balances exploration and exploitation.

\section{Implementation Details}
\label{sec:appendix-implementation-details}

We provide an overview of the implementation details of our method in Section \ref{sec:experiments}. For completeness, we list all relevant hyperparameters in Table \ref{tab:tdmpc-hparams}. As discussed in Appendix \ref{sec:appendix-baselines}, we adopt most hyperparameters from the SAC implementation \citep{pytorch_sac}. Following previous work \citep{hafner2019planet}, we use a task-specific action repeat hyperparameter for DMControl that is constant across all methods; see Table \ref{tab:action-repeat} for a list of values. For state-based experiments, we implement the representation function $h_{\theta}$ using an MLP with a single hidden layer of dimension $256$. For image-based experiments, $h_{\theta}$ is a 4-layer CNN with kernel sizes $(7,5,3,3)$, stride $(2,2,2,2)$, and $32$ filters per layer. All other components are implemented using 2-layer MLPs with dimension $512$. Following prior work \citep{pytorch_sac, srinivas2020curl, kostrikov2020image}, we apply layer normalization to the value function. Weights and biases in the last layer of the reward predictor $R_{\theta}$ and value function $Q_{\theta}$ are zero-initialized to reduce model and value biases in the early stages of training, and all other fully-connected layers use orthogonal initialization; the SAC and LOOP baselines are implemented similarly. We do not find it consistently better to use larger networks neither for state-based nor image-based experiments. In multi-task experiments, we augment the state input with a one-hot task vector. In multi-modal experiments, we encode state and image separately and sum the features. We provide a PyTorch-like summary of our task-oriented latent dynamics model in the following. For clarity, we use \emph{S}, \emph{Z}, and \emph{A} to denote the dimensionality of states, latent states, and actions, respectively, and report the total number of learnable parameters for our TOLD model initialized for the \textit{Walker Run} task ($\mathcal{S} \in \mathbb{R}^{24},~\mathcal{A} \in \mathbb{A}^{6}$).

{\small
\begin{codesnippet}
Total parameters: approx. 1,507,000
(h): Sequential(
  (0): Linear(in_features=S, out_features=256)
  (1): ELU(alpha=1.0)
  (2): Linear(in_features=256, out_features=Z))
(d): Sequential(
  (0): Linear(in_features=Z+A, out_features=512)
  (1): ELU(alpha=1.0)
  (2): Linear(in_features=512, out_features=512)
  (3): ELU(alpha=1.0)
  (4): Linear(in_features=512, out_features=Z))
(R): Sequential(
  (0): Linear(in_features=Z+A, out_features=512)
  (1): ELU(alpha=1.0)
  (2): Linear(in_features=512, out_features=512)
  (3): ELU(alpha=1.0)
  (4): Linear(in_features=512, out_features=1))
(pi): Sequential(
  (0): Linear(in_features=Z, out_features=512)
  (1): ELU(alpha=1.0)
  (2): Linear(in_features=512, out_features=512)
  (3): ELU(alpha=1.0)
  (4): Linear(in_features=512, out_features=A))
(Q1): Sequential(
  (0): Linear(in_features=Z+A, out_features=512)
  (1): LayerNorm((512,), elementwise_affine=True)
  (2): Tanh()
  (3): Linear(in_features=512, out_features=512)
  (4): ELU(alpha=1.0)
  (5): Linear(in_features=512, out_features=1))
(Q2): Sequential(
  (0): Linear(in_features=Z+A, out_features=512)
  (1): LayerNorm((512,), elementwise_affine=True)
  (2): Tanh()                                 
  (3): Linear(in_features=512, out_features=512)
  (4): ELU(alpha=1.0)
  (5): Linear(in_features=512, out_features=1))
\end{codesnippet}
}

Additionally, PyTorch-like pseudo-code for training our TOLD model (codified version of Algorithm \ref{alg:training}) is shown below:

{\small
\begin{codesnippet}
def update(replay_buffer):
  """
  A single gradient update of our TOLD model.
  h, R, Q, d: TOLD components.
  c1, c2, c3: loss coefficients.
  rho: temporal loss coefficient.
  """
  states, actions, rewards = replay_buffer.sample()
  
  # Encode first observation
  z = h(states[0])
  
  # Recurrently make predictions
  reward_loss = 0
  value_loss = 0
  consistency_loss = 0
  for t in range(H):
    r = R(z, actions[t])
    q1, q2 = Q(z, actions[t])
    z = d(z, actions[t])
    
    # Compute targets and losses
    z_target = h_target(states[t+1])
    td_target = compute_td(rewards[t], states[t+1])
    reward_loss += rho**t * mse(r, rewards[t])
    value_loss += rho**t * \
      (mse(q1, td_target) + mse(q2, td_target))
    consistency_loss += rho**t * mse(z, z_target)

  # Update
  total_loss = c1 * reward_loss + \
               c2 * value_loss + \
               c3 * consistency_loss
  total_loss.backward()
  optim.step()
 
  # Update slow-moving average
  update_target_network()
\end{codesnippet}
}

\begin{table}
\centering
\vspace{-0.225in}
\parbox{0.48\textwidth}{
\caption{\textbf{TD-MPC hyperparameters.} We here list hyperparameters for TD-MPC with TOLD and emphasize that we use the same parameters for SAC whenever possible.}
\label{tab:tdmpc-hparams}
\vspace{0.05in}
\centering
\resizebox{0.45\textwidth}{!}{%
\begin{tabular}{@{}ll@{}}
\toprule
Hyperparameter                                                                   & Value                                                                             \\ \midrule
Discount factor ($\gamma$)                                                         & 0.99                                                                              \\
Seed steps                                                                  & $5,000$                                                                          \\
Replay buffer size                                                        & Unlimited                                                                                      \\
Sampling technique                                                        & PER ($\alpha=0.6, \beta=0.4$)                                                                                      \\
Planning horizon ($H$)                                                         & $5$                                                                              \\
Initial parameters ($\mu^{0}, \sigma^{0}$)                                 & $(0, 2)$                                                                          \\
Population size                                                            & $512$                                                                          \\
Elite fraction                                                             & $64$                                                                          \\
\begin{tabular}[c]{@{}l@{}}Iterations\\~\\~ \end{tabular}                                                                    & \begin{tabular}[c]{@{}l@{}}12 (Humanoid)\\8 (Dog, pixels)\\6 (otherwise)\end{tabular}\\ 
Policy fraction                                                            & $5\%$                                                                          \\
Number of particles                                                        & $1$                                                                          \\
Momentum coefficient                                                       & $0.1$                                                                          \\
Temperature ($\tau$)                                                 & $0.5$                                                                         \\
MLP hidden size                                                                & $512$                                                                          \\
MLP activation                                                                 & ELU
                                        \\
\begin{tabular}[c]{@{}l@{}}Latent dimension\\~ \end{tabular}                                                                    & \begin{tabular}[c]{@{}l@{}}100 (Humanoid, Dog)\\50 (otherwise)\end{tabular}\\ 
\begin{tabular}[c]{@{}l@{}}Learning rate\\~ \end{tabular}                                                                    & \begin{tabular}[c]{@{}l@{}}3e-4 (Dog, pixels)\\1e-3 (otherwise)\end{tabular}\\ 
Optimizer ($\theta$)                                                & Adam ($\beta_1=0.9, \beta_2=0.999$)                                                       \\
Temporal coefficient ($\lambda$)                                              & $0.5$                                                                          \\
Reward loss coefficient ($c_{1}$)                                              & $0.5$                                                                          \\
Value loss coefficient ($c_{2}$)                                              & $0.1$                                                                          \\
Consistency loss coefficient ($c_{3}$)                                        & $2$                                                                          \\
Exploration schedule ($\epsilon$)                                             & $0.5\rightarrow 0.05$ (25k steps)                                                                          \\
Planning horizon schedule                                             & $1\rightarrow 5$ (25k steps)                                                                          \\
\begin{tabular}[c]{@{}l@{}}Batch size\\~\\~ \end{tabular}                                                                    & \begin{tabular}[c]{@{}l@{}}2048 (Dog)\\256 (pixels)\\512 (otherwise)\end{tabular}\\ 
Momentum coefficient ($\zeta$)                                                 & $0.99$                                                                         \\
Steps per gradient update                                                     & $1$                                                                          \\
$\theta^{-}$ update frequency                                                  & 2                                                                 \\ \bottomrule
\end{tabular}%
}
}
\vspace{-0.15in}
\end{table}

\begin{table}
\centering
\parbox{0.48\textwidth}{
\caption{\textbf{SAC hyperparameters.} We list the most important hyperparameters for the SAC baseline. Note that we mostly follow the implementation of \citet{pytorch_sac} but improve upon certain hyperparameter choices, e.g., the momentum coefficient $\zeta$ and values specific to the Dog tasks.}
\label{tab:sac-hparams}
\vspace{0.05in}
\centering
\resizebox{0.41\textwidth}{!}{%
\begin{tabular}{@{}ll@{}}
\toprule
Hyperparameter                                                                   & Value                                                                             \\ \midrule
Discount factor ($\gamma$)                                                         & 0.99                                                                              \\
Seed steps                                                                  & $1,000$                                                                          \\
Replay buffer size                                                        & Unlimited                                                                                      \\
Sampling technique                                                        & Uniform                                                                                      \\
MLP hidden size                                                                & $1024$                                                                          \\
MLP activation                                                                 & RELU
                                        \\
\begin{tabular}[c]{@{}l@{}}Latent dimension\\~ \end{tabular}                                                                    & \begin{tabular}[c]{@{}l@{}}100 (Humanoid, Dog)\\50 (otherwise)\end{tabular}\\ 
Optimizer ($\theta$)                                                & Adam ($\beta_1=0.9, \beta_2=0.999$)                                                       \\
Optimizer ($\alpha$ of SAC)                                                      & Adam ($\beta_1=0.5, \beta_2=0.999$)                                                       \\
\begin{tabular}[c]{@{}l@{}}Learning rate ($\theta$)  \\~ \end{tabular}                                                                    & \begin{tabular}[c]{@{}l@{}}3e-4 (Dog)\\1e-3 (otherwise)\end{tabular}\\ 
Learning rate ($\alpha$ of SAC)                                                  & 1e-4                                                                              \\
\begin{tabular}[c]{@{}l@{}}Batch size\\~ \end{tabular}                                                                    & \begin{tabular}[c]{@{}l@{}}2048 (Dog)\\512 (otherwise)\end{tabular}\\
Momentum coefficient ($\zeta$)                                                & 0.99 \\
Steps per gradient update                                               & 1 \\
$\theta^{-}$ update frequency                                                  & 2                                                                 \\ \bottomrule
\end{tabular}%
}
}
\vspace{-0.1in}
\end{table}

\section{Extended Description of Baselines}
\label{sec:appendix-baselines}

We tune the performance of both our method and baselines to perform well on DMControl and then subsequently benchmark algorithms on Meta-World using the same choice of hyperparameters. Below, we provide additional details on our efforts to tune the baseline implementations.

\textbf{SAC.} We adopt the implementation of \citet{pytorch_sac} which has been used extensively in the literature as a benchmark implementation for state-based DMControl. We use original hyperparameters except for the target network momentum coefficient $\zeta$, where we find it beneficial for both SAC, LOOP, and our method to use a faster update of $\zeta = 0.99$ as opposed to $0.995$ in the original implementation. Additionally, we decrease the batch size from $1024$ to $512$ for fair comparison to our method. For completeness, we list important hyperparameters for the SAC baseline in Table \ref{tab:sac-hparams}.

\textbf{LOOP.} We benchmark against the official implementation from \citet{Sikchi2020LearningOW}, but note that LOOP has -- to the best of our knowledge -- not previously been benchmarked on DMControl nor Meta-World. Therefore, we do our best to adapt its hyperparameters. As in the SAC implementation, we find LOOP to perform better using $\zeta = 0.99$ than its original value of $0.995$, and we increase the batch size from $256$ to $512$. Lastly, we set the number of seed steps to $1,000$ (down from $10,000$) to match the SAC implementation. As LOOP uses SAC as backbone learning algorithm, we found these changes to be beneficial. LOOP-specific hyperparameters are listed in Table \ref{tab:loop-hparams}.

\textbf{MPC:sim.} We compare TD-MPC to a vanilla MPC algorithm using a ground-truth model of the environment (simulator), but no terminal value function. As such, this baseline is non-parametric. We use the same MPC implementation as in our method (MPPI; \citet{Williams2015ModelPP}). As planning with a simulator is computationally intensive, we limit the planning horizon to $10$ (which is still $2\times$ as much as TD-MPC), and we reduce the number of iterations to 4 (our method uses 6), as we find MPC to converge faster when using the ground-truth model. At each iteration, we sample $N=200$ trajectories and update distribution parameters using the top-$20$ ($10\%$) sampled action sequences. We keep all other hyperparameters consistent with our method. Because of the limited planning horizon, this MPC baseline generally performs well for locomotion tasks where local solutions are sufficient, but tends to fail at tasks with, for example, sparse rewards.

\textbf{No latent ablation.} We make the following change to our method: replacing $h_{\theta}$ with the identity function, i.e., $\mathbf{x} = h_{\theta}(\mathbf{x})$. As such, environment dynamics are modelled by forward prediction directly in the state space, with the consistency loss effectively degraded to a state prediction loss. This ablation makes our method more similar to prior work on model-based RL from states \citep{Janner2019WhenTT, Lowrey2019PlanOL, Sikchi2020LearningOW, Argenson2021ModelBasedOP}. However, unlike previous work that decouples model learning from policy and value learning, we still back-propagate gradients from the reward and value objectives through the model, which is a stronger baseline.

\textbf{No consistency regularization.} We set the coefficient $c_{3}$ corresponding to the latent state consistency loss in Equation \ref{eq:consistency-loss} to $0$, such that the TOLD model is trained only with the reward and value prediction losses. This ablation makes our method more similar to MuZero \citep{Schrittwieser2020MasteringAG}.

\textbf{Other baselines.} Results for other baselines are obtained from related work. Specifically, results for SAC, CURL, DrQ, and PlaNet are obtained from \citet{srinivas2020curl} and \citet{kostrikov2020image}, and results for Dreamer, MuZero, and EfficientZero are obtained from \citet{Hafner2020DreamTC} and \citet{Ye2021MasteringAG}.

\begin{table}
\centering
\parbox{0.48\textwidth}{
\caption{\textbf{LOOP hyperparameters.} We list general SAC hyperparameters shared by LOOP in Table \ref{tab:sac-hparams}, and list only hyperparameters specific to LOOP here. We use the official implementation from \citet{Sikchi2020LearningOW} but list its hyperparameters for completeness. Note that we -- as in the SAC implementation -- use a different batch size and momentum coefficient than in \citet{Sikchi2020LearningOW}, as we find this to marginally improve performance on DMControl.}
\label{tab:loop-hparams}
\vspace{0.1in}
\centering
\resizebox{0.265\textwidth}{!}{%
\begin{tabular}{@{}ll@{}}
\toprule
Hyperparameter                                                                   & Value                                                                             \\ \midrule
Planning horizon ($H$)                                                         & 3                                                                              \\
Population size                                                            & $100$                                                                          \\
Elite fraction                                                             & $20\%$                                                                          \\
Iterations                                                                 & $5$                                                                          \\
Policy fraction                                                            & $5\%$                                                                          \\
Number of particles                                                        & $4$                                                                          \\
Momentum coefficient                                                       & $0.1$                                                                          \\
MLP hidden size                                                                & $256$                                                                          \\
MLP activation                                                                 & ELU/RELU
                                        \\
Ensemble size                                                                  & $5$                                                                          \\ \bottomrule
\end{tabular}%
}
}
\vspace{-0.1in}
\end{table}

\begin{table}
\centering
\parbox{0.48\textwidth}{
\caption{\textbf{Action repeat.} We adopt action repeat hyperparameters for DMControl from previous work \citep{hafner2019planet, kostrikov2020image} for state-based experiments as well as the DMControl 100k benchmark; we list all values below. For the DMControl \textit{Dreamer} benchmark, all methods use an action repeat of 2 regardless of the task. We do not use action repeat for Meta-World.}
\label{tab:action-repeat}
\vspace{0.1in}
\centering
\resizebox{0.26\textwidth}{!}{%
\begin{tabular}{@{}ll@{}}
\toprule
Task                                                                   & Action repeat                                                                             \\ \midrule
Humanoid                                                         & 2                                                                              \\
Dog                                                              & 2                                                                              \\
Walker                                                           & 2                                                                              \\
Finger                                                           & 2                                                                              \\
Cartpole                                                         & 8                                                                              \\
Other (DMControl)                                                & 4                                                                              \\
Meta-World                                                       & 1                                                                              \\ \bottomrule
\end{tabular}%
}
}
\vspace{-0.1in}
\end{table}

\begin{figure}
    \centering
    \includegraphics[width=0.48\textwidth]{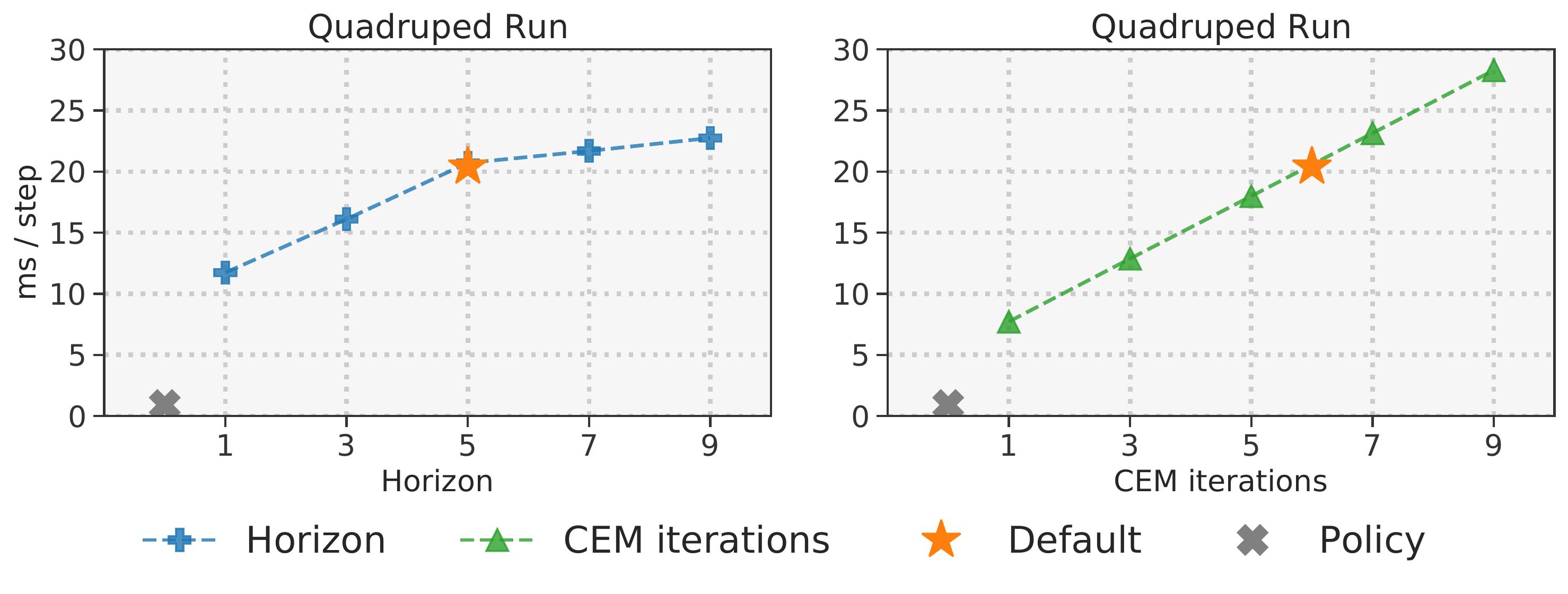}
    \vspace{-0.3in}
    \caption{\textbf{Inference time under a variable budget.} Milliseconds per decision step for TD-MPC on the \textit{Quadruped Run} task under a variable computational budget. We evaluate performance of fully trained agents when varying \textit{(left)} planning horizon; \textit{(right)} number of iterations during planning. When varying one hyperparameter, the other is fixed to the default value. For completeness, we also include the inference time of the learned policy $\pi_{\theta}$, and the default setting of $6$ iterations and a horizon of $5$ used during training.} 
    \label{fig:inference}
\end{figure}

\begin{table}
\vspace{-0.05in}
\caption{\textbf{Meta-World MT10.} As our performance metric reported in Figure \ref{fig:metaworld-all} differs from that of the Meta-World \textit{v2} benchmark proposal \citep{yu2019meta}, we here report results for our SAC baseline using the same \textit{maximum per-task success rate} metric used for the MT10 multi-task experiment from the original paper.}
\label{tab:appendix-mt10}
\vspace{0.075in}
\centering
\resizebox{0.3\textwidth}{!}{%
\begin{tabular}{lc}
\toprule
\textit{Task} & Max. success rate \\\midrule
Window Close      & $1.00$ \\
Window Open       & $1.00$ \\
Door Open         & $1.00$ \\
Peg Insert Side   & $0.00$ \\
Drawer Open       & $0.85$ \\
Pick Place        & $0.00$ \\
Reach             & $1.00$ \\
Button Press Down & $1.00$ \\
Push              & $0.00$ \\
Drawer Close      & $1.00$ \\
\bottomrule
\end{tabular}
}
\vspace{-0.025in}
\end{table}

\section{Inference Time}
\label{sec:appendix-inference}
In the experiments of Section \ref{sec:experiments}, we investigate the relationship between performance and the computational budget of planning with TD-MPC. For completeness, we also evaluate the relationship between computational budget and inference time. Figure \ref{fig:inference} shows the inference time of TD-MPC as the planning horizon and number of iterations is varied. As in previous experiments, we benchmark inference times on a single RTX3090 GPU. Unsurprisingly, we find that there is an approximately linear relationship between computational budget and inference time. However, it is worth noting that our default settings used during training only require approximately 20ms per step, i.e., 50Hz, which is fast enough for many real-time robotics applications such as manipulation, navigation, and to some extent locomotion (assuming an on-board GPU). For applications where inference time is critical, the computational budget can be adjusted to meet requirements. For example, we found in Figure \ref{fig:appendix-computational-budget} that we can reduce the planning horizon of TD-MPC on the \textit{Quadruped Run} task from 5 to 1 with no significant reduction in performance, which reduces inference time to approximately 12ms per step. While the performance of the model-free policy learned jointly with TD-MPC indeed is lower than that of planning, it is however still nearly $6\times$ faster than planning at inference time.

\section{Meta-World}
\label{sec:appendix-metaworld}
\begin{figure*}[h]
    \centering
    \includegraphics[width=0.9\textwidth]{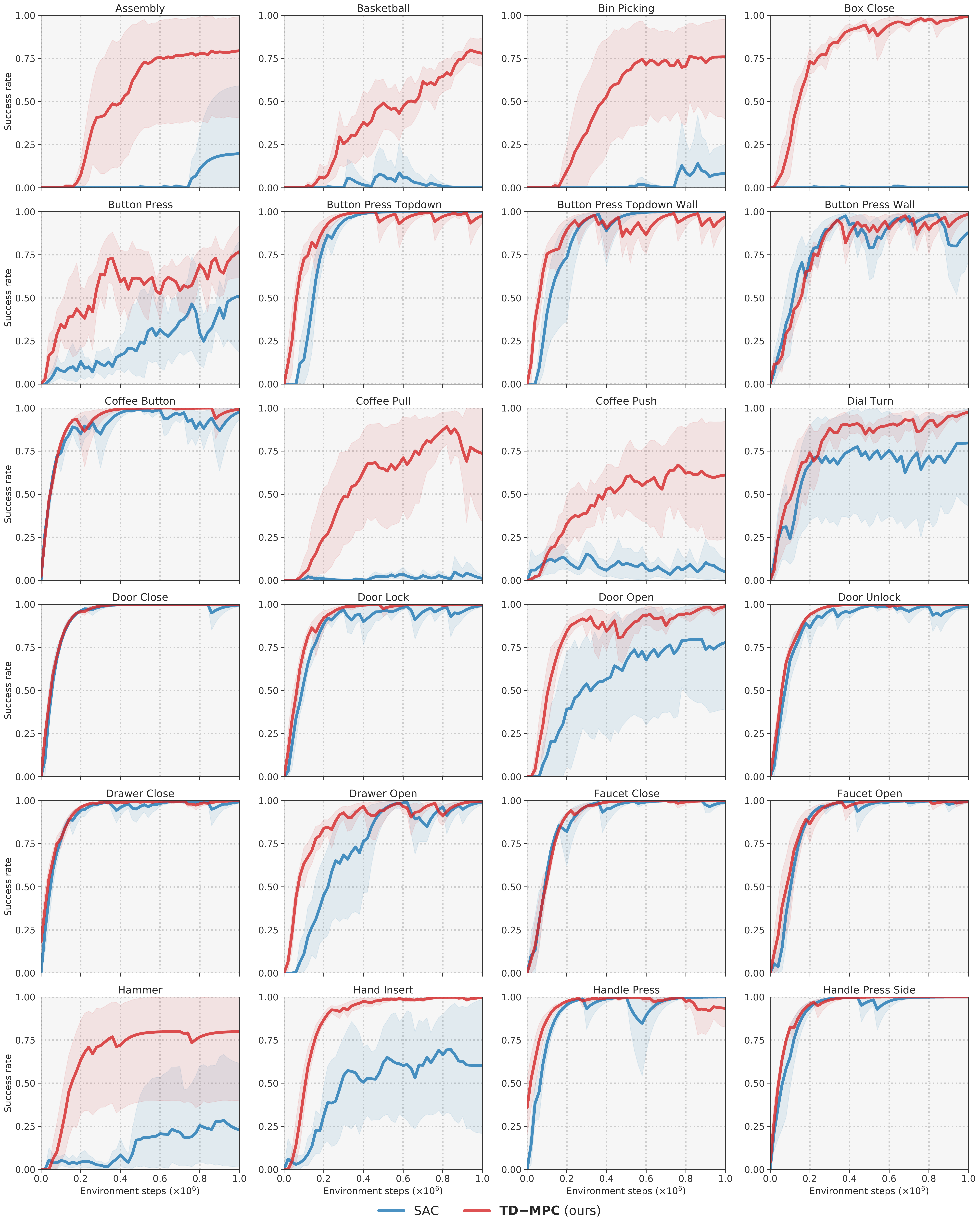}
    \vspace{-0.15in}
    \caption{\textbf{Individual Meta-World tasks.} Success rate of our method (TD-MPC) and SAC on diverse manipulation tasks from Meta-World \citep{yu2019meta}. We use the goal-conditioned version of Meta-World, which is considered harder than the fixed-goal version. Due to the large number of tasks (50), we choose to visualize only the first 24 tasks (sorted alphabetically). Mean of 5 runs; shaded areas are $95\%$ confidence intervals. Our method is capable of solving complex tasks (e.g., \textit{Basketball}) where SAC achieves a relatively small success rate. Note that we use Meta-World \textit{v2} and performances are therefore not comparable to previous work using \textit{v1}.}
    \label{fig:meta-all}
    \vspace{-0.2in}
\end{figure*}

We provide learning curves and success rates for individual Meta-World \citep{yu2019meta} tasks in Figure \ref{fig:meta-all}. Due to the sheer number of tasks, we choose to only visualize the first 24 tasks (sorted alphabetically) out of the total of 50 tasks from Meta-World. Note that we use Meta-World \textit{v2} and that we consider the goal-conditioned versions of the tasks, which are considered harder than the single-goal variant often used in related work. We generally find that SAC is competitive to TD-MPC in most tasks, but that TD-MPC is far more sample efficient in tasks that involve complex manipulation, e.g., \textit{Bin Picking}, \textit{Box Close}, and \textit{Hammer}. Successful trajectories for each of these three tasks are visualized in Figure \ref{fig:appendix-visualizations}. Generally, we choose to focus on sample-efficiency for which we empirically find 1M environment steps (3M for multi-task experiments) to be sufficient for achieving non-trivial success rates in Meta-World. As the original paper reports \textit{maximum per-task success rate} for multi-task experiments rather than average success rate, we also report this metric for our SAC baseline in Table \ref{tab:appendix-mt10}. We find that our SAC baseline is strikingly competitive with the original paper results considering that we evaluate over just 3M steps.

\begin{figure*}[t]
    \centering
    \includegraphics[width=0.85\textwidth]{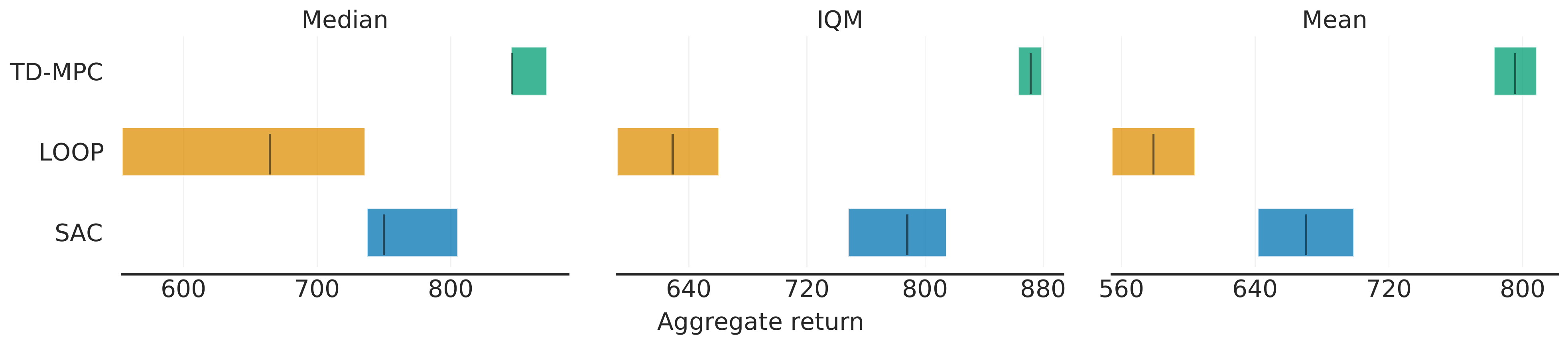}
    \vspace{-0.1in}
    \caption{\textbf{Rliable metrics.} Median, interquantile median (IQM), and mean performance of TD-MPC and baselines on the 15 state-based DMControl tasks. Confidence intervals are estimated using the percentile bootstrap with stratified sampling, per recommendation of \citet{agarwal2021deep}. Higher values are better. 5 seeds.}
    \label{fig:rliable}
    \vspace{-0.05in}
\end{figure*}

\section{Multi-Modal RL}
\label{sec:appendix-multimodal}

We demonstrate the ability of TD-MPC to successfully fuse information from multiple input modalities (proprioceptive data + an egocentric camera) in two 3D locomotion tasks:

$\boldsymbol{-}$ \textbf{Quadruped Corridor}, where the agent needs to move along a corridor with constant target velocity. To succeed, the agent must perceive the corridor walls and adjust its walking direction accordingly. \vspace{0.05in}\\
$\boldsymbol{-}$ \textbf{Quadruped Obstacles}, where the agent needs to move along a corridor filled with obstacles that obstruct vision and forces the agent to move in a zig-zag pattern with constant target velocity. To succeed, the agent must perceive both the corridor walls and obstacles, and continuously adjust its walking direction.

\begin{figure}
    \centering
    \rotatebox{90}{\footnotesize~~~~~~Corridor}~
    \includegraphics[width=0.125\textwidth]{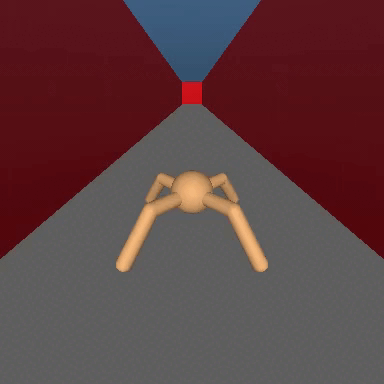}
    \includegraphics[width=0.125\textwidth]{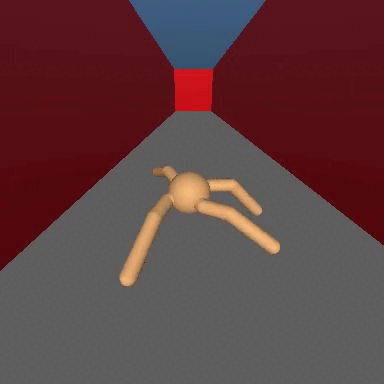}
    \includegraphics[width=0.125\textwidth]{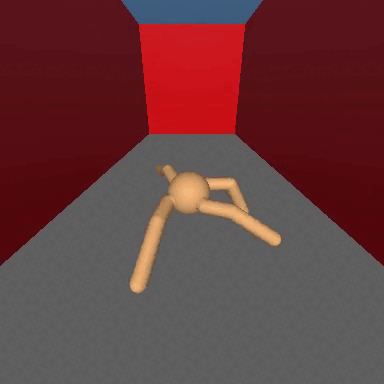}\vspace{0.05in}\\
    ~\rotatebox{90}{\footnotesize~~~~~Obstacles}~
    \includegraphics[width=0.125\textwidth]{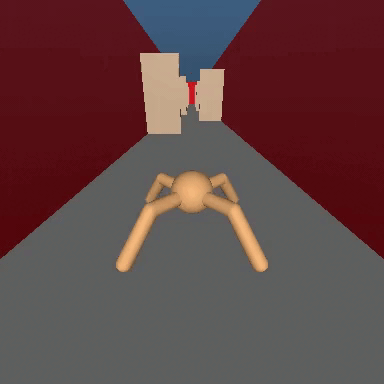}
    \includegraphics[width=0.125\textwidth]{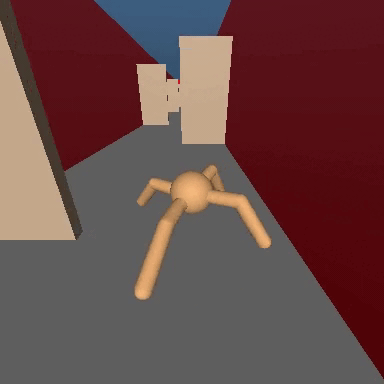}
    \includegraphics[width=0.125\textwidth]{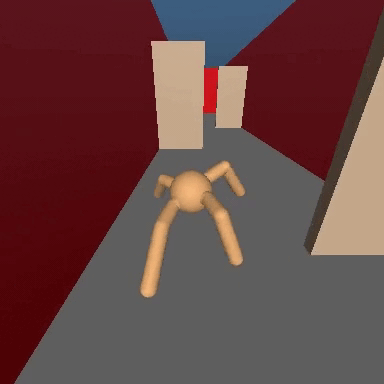}
    \vspace{-0.1in}
    \caption{\textbf{Multi-modal RL.} Visualization of the two multi-modal 3D locomotion tasks that we construct.}
    \label{fig:multimodal}
    \vspace{-0.1in}
\end{figure}

Trajectories from the two tasks are visualized in Figure \ref{fig:multimodal}.

\section{Additional Metrics}
\label{sec:appendix-rliable}

We report additional (aggregate) performance metrics of SAC, LOOP, and TD-MPC on the set of 15 state-based DMControl tasks using the \textit{rliable} toolkit provided by \citet{agarwal2021deep}. Concretely, we report the aggregate median, interquantile mean (IQM), and mean returns with 95\% confidence intervals based on the episode returns of trained (after 500k environment steps) agents. As recommended by \citet{agarwal2021deep}, confidence intervals are estimated using the percentile bootstrap with stratified sampling.

\section{Task Visualizations}
\label{sec:appendix-visualizations}

\begin{figure*}[t]
    \centering
    time $\longrightarrow$\vspace{0.05in}\\
    \rotatebox{90}{\footnotesize~~~~~~~~Dog Walk}~
    \includegraphics[width=0.155\textwidth]{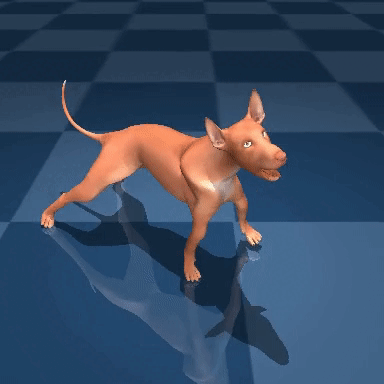}
    \includegraphics[width=0.155\textwidth]{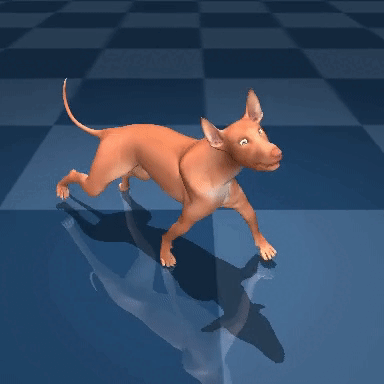}
    \includegraphics[width=0.155\textwidth]{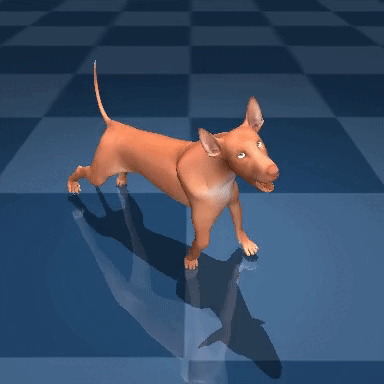}
    \includegraphics[width=0.155\textwidth]{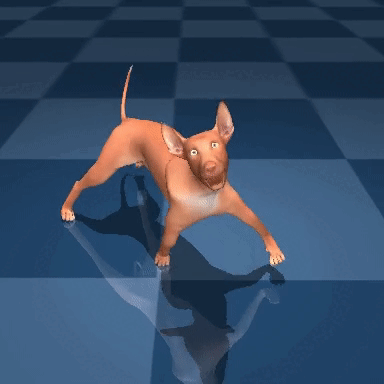}
    \includegraphics[width=0.155\textwidth]{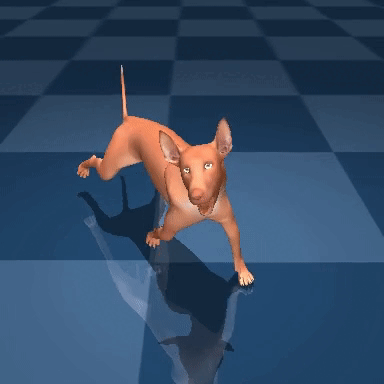}
    \includegraphics[width=0.155\textwidth]{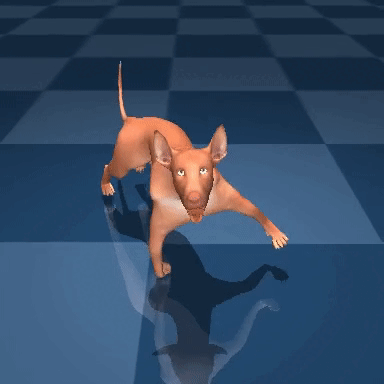}\vspace{0.05in}\\
    ~\rotatebox{90}{\footnotesize~~~Humanoid Walk}~
    \includegraphics[width=0.155\textwidth]{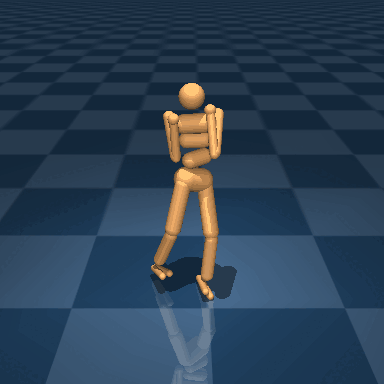}
    \includegraphics[width=0.155\textwidth]{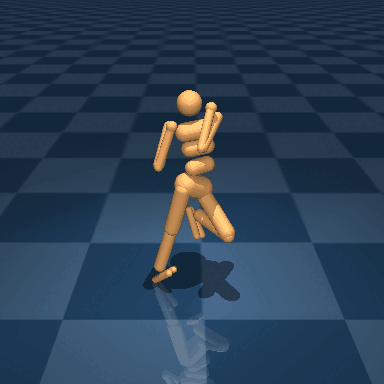}
    \includegraphics[width=0.155\textwidth]{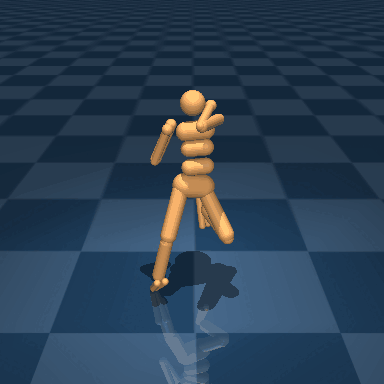}
    \includegraphics[width=0.155\textwidth]{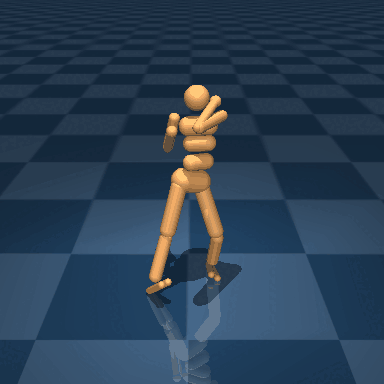}
    \includegraphics[width=0.155\textwidth]{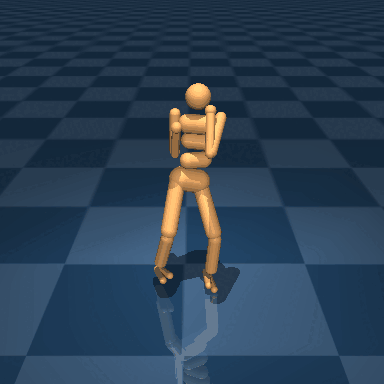}
    \includegraphics[width=0.155\textwidth]{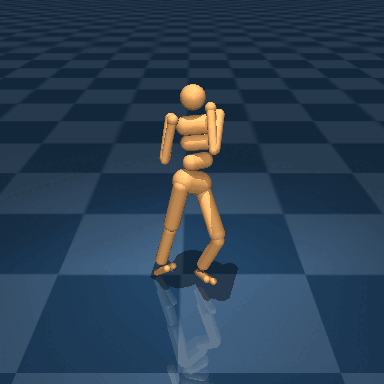}\vspace{0.05in}\\
    ~\rotatebox{90}{\footnotesize~~~~Quadruped Run}~
    \includegraphics[width=0.155\textwidth]{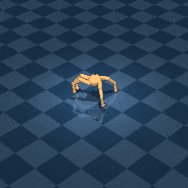}
    \includegraphics[width=0.155\textwidth]{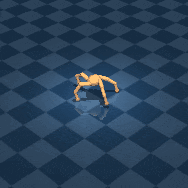}
    \includegraphics[width=0.155\textwidth]{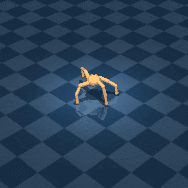}
    \includegraphics[width=0.155\textwidth]{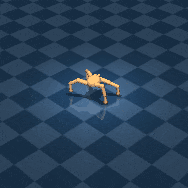}
    \includegraphics[width=0.155\textwidth]{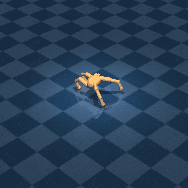}
    \includegraphics[width=0.155\textwidth]{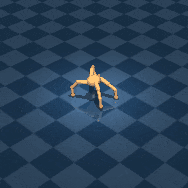}\vspace{0.05in}\\
    ~\rotatebox{90}{\footnotesize~~~Finger Turn Hard}~
    \includegraphics[width=0.155\textwidth]{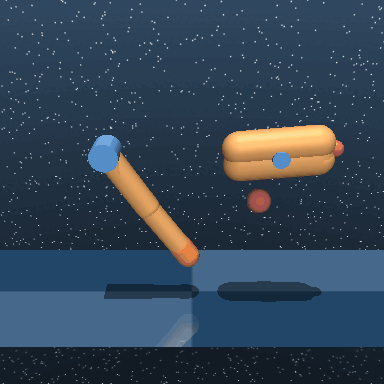}
    \includegraphics[width=0.155\textwidth]{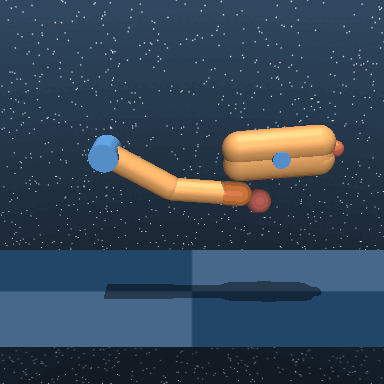}
    \includegraphics[width=0.155\textwidth]{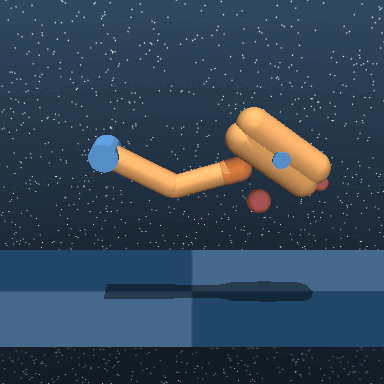}
    \includegraphics[width=0.155\textwidth]{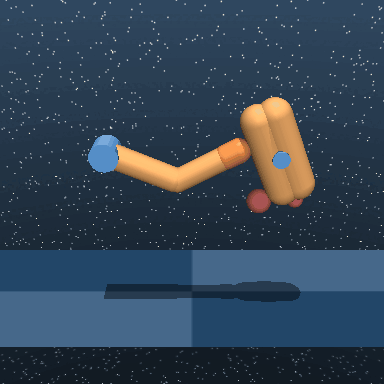}
    \includegraphics[width=0.155\textwidth]{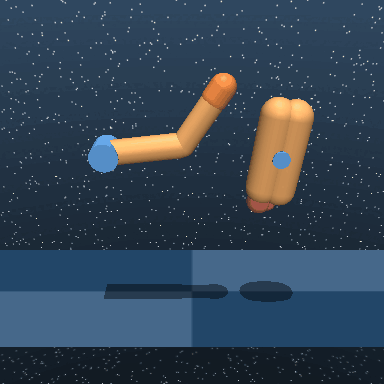}
    \includegraphics[width=0.155\textwidth]{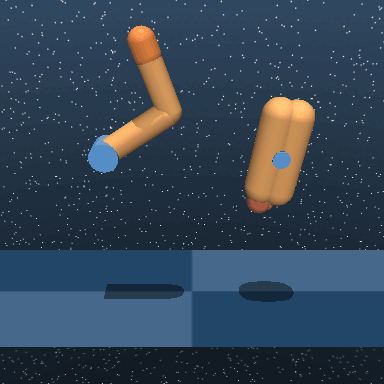}\vspace{0.05in}\\
    ~\rotatebox{90}{\footnotesize~~~~~~Bin Picking}~
    \includegraphics[width=0.155\textwidth]{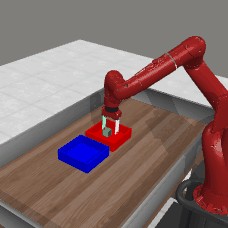}
    \includegraphics[width=0.155\textwidth]{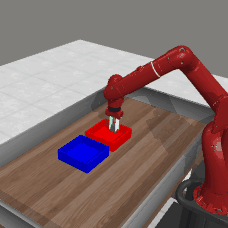}
    \includegraphics[width=0.155\textwidth]{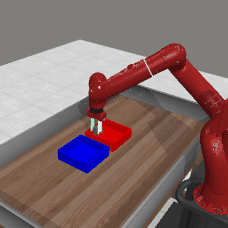}
    \includegraphics[width=0.155\textwidth]{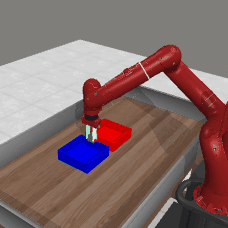}
    \includegraphics[width=0.155\textwidth]{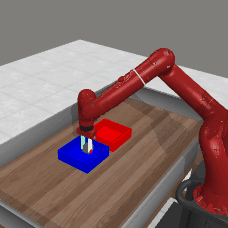}
    \includegraphics[width=0.155\textwidth]{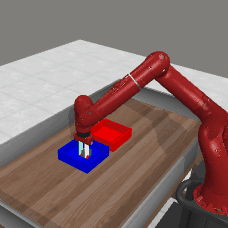}\vspace{0.05in}\\
    ~\rotatebox{90}{\footnotesize~~~~~~~~Box Close}~
    \includegraphics[width=0.155\textwidth]{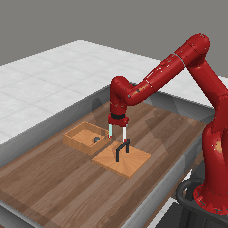}
    \includegraphics[width=0.155\textwidth]{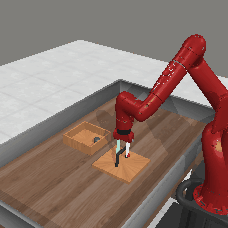}
    \includegraphics[width=0.155\textwidth]{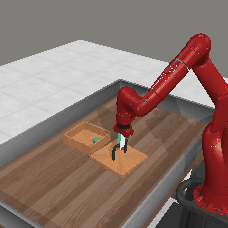}
    \includegraphics[width=0.155\textwidth]{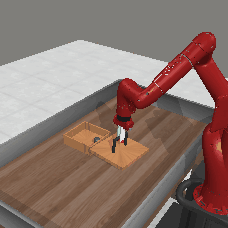}
    \includegraphics[width=0.155\textwidth]{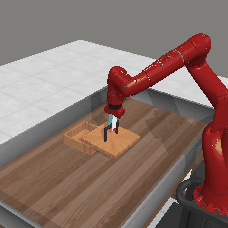}
    \includegraphics[width=0.155\textwidth]{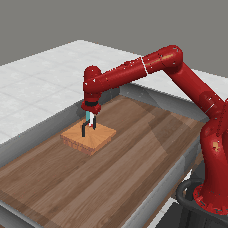}\vspace{0.05in}\\
    ~\rotatebox{90}{\footnotesize~~~~~~~~~Hammer}~
    \includegraphics[width=0.155\textwidth]{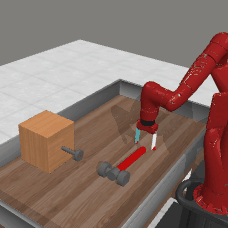}
    \includegraphics[width=0.155\textwidth]{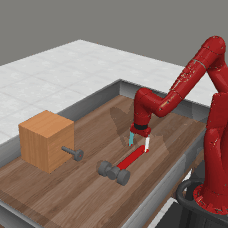}
    \includegraphics[width=0.155\textwidth]{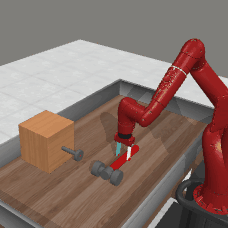}
    \includegraphics[width=0.155\textwidth]{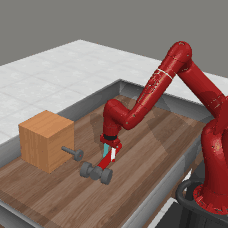}
    \includegraphics[width=0.155\textwidth]{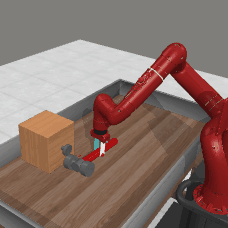}
    \includegraphics[width=0.155\textwidth]{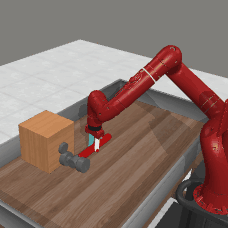}\\
    \vspace{-0.05in}
    \caption{\textbf{Visualizations.} We visualize trajectories generated by our method on seven selected tasks from the two benchmarks, listed (from top to bottom) as follows: \textit{(1)} Dog Walk, a challenging locomotion task that has a high-dimensional action space ($\mathcal{A} \in \mathbb{R}^{38}$); \textit{(2)} Humanoid Walk, a challenging locomotion task ($\mathcal{A} \in \mathbb{R}^{21}$); \textit{(3)} Quadruped Run, a four-legged locomotion task ($\mathcal{A} \in \mathbb{R}^{12}$); \textit{(4)} Finger Turn Hard, a hard exploration task with sparse rewards; \textit{(5)} Bin Picking, a 3-d pick-and-place task; \textit{(6)} Box Close, a 3-d manipulation task; and lastly \textit{(7)} Hammer, another 3d-manipulation task. In all seven trajectories, we display only key frames in the trajectory. Actual episode lengths are 1000 (DMControl) and 500 (Meta-World). Our method (TD-MPC) is capable of solving each of these tasks in less than 1M environment steps. Video results are available at \url{https://nicklashansen.github.io/td-mpc}.}
    \label{fig:appendix-visualizations}
    \vspace{-0.1in}
\end{figure*}

Figure \ref{fig:appendix-visualizations} provides visualizations of successful trajectories generated by TD-MPC on seven tasks from DMControl and Meta-World, all of which TD-MPC solves in less than 1M environment steps. In all seven trajectories, we display only key frames in the trajectory, as actual episode lengths are 1000 (DMControl) and 500 (Meta-World). For full video trajectories, refer to \url{https://nicklashansen.github.io/td-mpc}.

\textbf{Additional material on the following pages $\downarrow$}

\end{document}

%% file: tables/planet.tex
\begin{table*}[t!]
\caption{\textbf{Learning from pixels.} Return of our method (TD-MPC) and state-of-the-art algorithms on the image-based DMControl 100k benchmark used in \citet{srinivas2020curl, kostrikov2020image, Ye2021MasteringAG}. Baselines are tuned specifically for image-based RL, whereas our method is not. Results for SAC, CURL, DrQ, and PlaNet are partially obtained from \citet{srinivas2020curl, kostrikov2020image}, and results for Dreamer, MuZero, and EfficientZero are obtained from \citet{Hafner2020DreamTC, Ye2021MasteringAG}. Mean and std. deviation over 10 runs. \textbf{*}: MuZero and EfficientZero use a discretized action space, and EfficientZero performs an additional 20k gradient steps before evaluation, whereas other methods do not. Due to dimensionality explosion under discretization, MuZero and EfficientZero cannot feasibly solve tasks with higher-dimensional action spaces, e.g., Walker Walk and Cheetah Run ($\mathcal{A} \in \mathbb{R}^{6}$), while our method can.}
\label{tab:planet}
\vspace{0.05in}
\centering
\resizebox{0.91\textwidth}{!}{%
\begin{tabular}{lcccc|cccc|c}
\multicolumn{1}{c}{} & \multicolumn{4}{c}{Model-free} & \multicolumn{4}{c}{Model-based} & \multicolumn{1}{c}{Ours} \\
\toprule
\textit{100k env. steps} & SAC State & SAC Pixels & CURL & DrQ & PlaNet & Dreamer & MuZero\textbf{*} & Eff.Zero\textbf{*} & \textbf{TD-MPC} \\\midrule
Cartpole Swingup  & $812 \scriptstyle{\pm 45}$ & $419 \scriptstyle{\pm 40}$ & $597 \scriptstyle{\pm 170}$ & $\mathbf{759 \scriptstyle{\pm 92}}$ & $563 \scriptstyle{\pm 73}$ & $326 \scriptstyle{\pm 27}$ & $219 \scriptstyle{\pm 122}$ & $\mathbf{813\scriptstyle{\pm 19}}$ & $\mathbf{770\scriptstyle{\pm 70}}$ \\
Reacher Easy      & $919 \scriptstyle{\pm 123}$ & $145 \scriptstyle{\pm 30}$ & $517 \scriptstyle{\pm 113}$ & $601 \scriptstyle{\pm 213}$ & $82 \scriptstyle{\pm 174}$ & $314 \scriptstyle{\pm 155}$ & $493 \scriptstyle{\pm 145}$ & $\mathbf{952\scriptstyle{\pm 34}}$ & $628\scriptstyle{\pm 105}$ \\
Cup Catch         & $957 \scriptstyle{\pm 26}$ & $312 \scriptstyle{\pm 63}$ & $772 \scriptstyle{\pm 241}$ & $\mathbf{913 \scriptstyle{\pm 53}}$ & $710 \scriptstyle{\pm 217}$ & $246 \scriptstyle{\pm 174}$ & $542 \scriptstyle{\pm 270}$ & $\mathbf{942\scriptstyle{\pm 17}}$ & $\mathbf{933\scriptstyle{\pm 24}}$ \\
Finger Spin       & $672 \scriptstyle{\pm 76}$ & $166 \scriptstyle{\pm 128}$ & $779 \scriptstyle{\pm 108}$ & $\mathbf{901 \scriptstyle{\pm 104}}$ & $560 \scriptstyle{\pm 77}$ & $341 \scriptstyle{\pm 70}$ & $-$ & $-$ & $\mathbf{943 \scriptstyle{\pm 59}}$ \\
Walker Walk       & $604 \scriptstyle{\pm 317}$ & $42 \scriptstyle{\pm 12}$ & $344 \scriptstyle{\pm 132}$ & $\mathbf{612 \scriptstyle{\pm 164}}$ & $221 \scriptstyle{\pm 43}$ & $277 \scriptstyle{\pm 12}$ & $-$ & $-$ & $\mathbf{577 \scriptstyle{\pm 208}}$ \\
Cheetah Run       & $228 \scriptstyle{\pm 95}$ & $103 \scriptstyle{\pm 38}$ & $\mathbf{307 \scriptstyle{\pm 48}}$ & $\mathbf{344 \scriptstyle{\pm 67}}$ & $165 \scriptstyle{\pm 123}$ & $235 \scriptstyle{\pm 137}$ & $-$ & $-$ & $222 \scriptstyle{\pm 88}$\\
\bottomrule
\end{tabular}
}
\end{table*}
\begin{figure*}[t!]
    \centering
    \vspace{-0.05in}
    \includegraphics[width=0.88\textwidth]{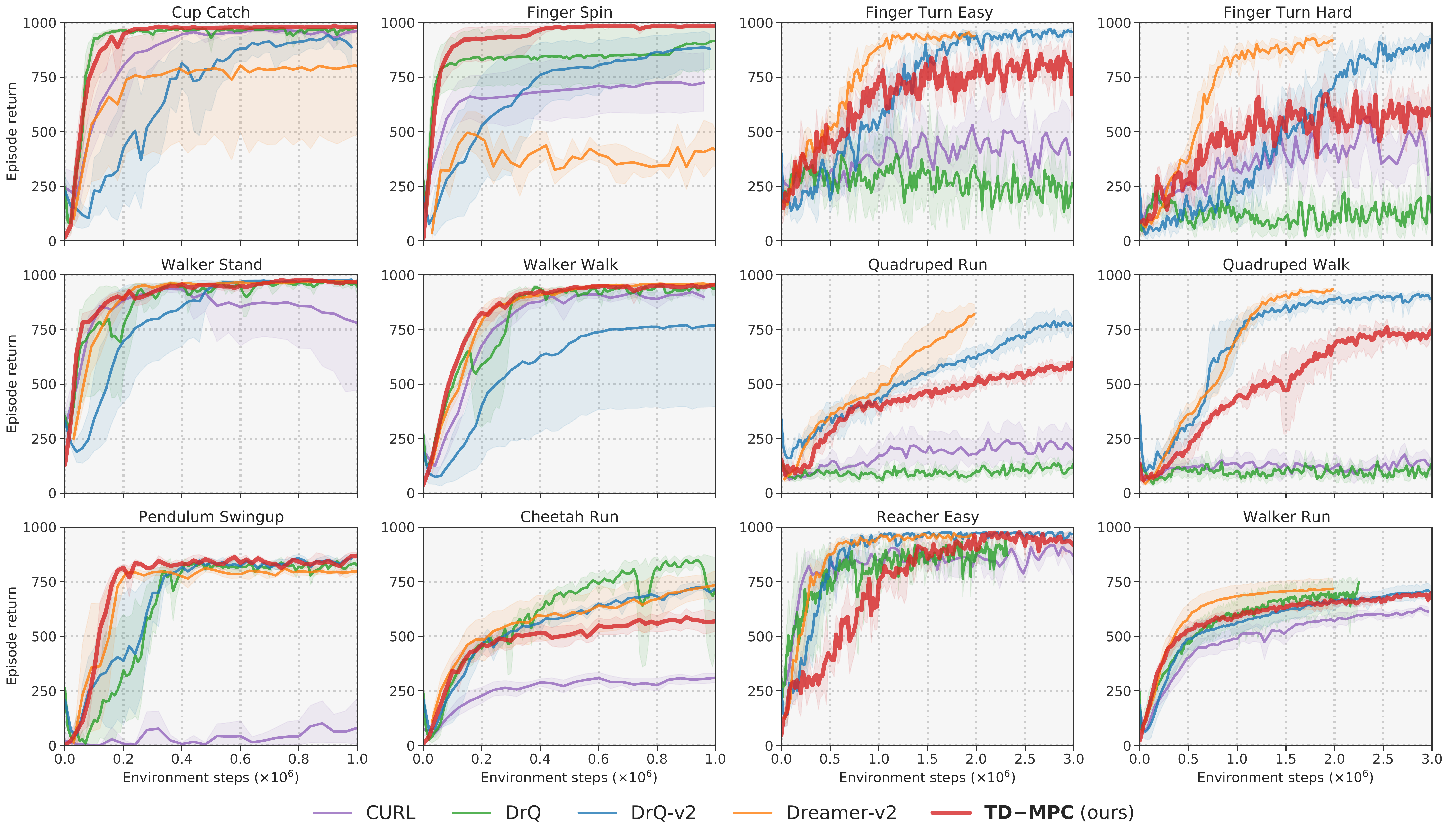}
    \vspace{-0.2in}
    \caption{\textbf{Learning from pixels.} Return of our method (TD-MPC) and state-of-the-art algorithms on 12 challenging image-based DMControl tasks. We follow prior work \citep{Hafner2020DreamTC, hafner2020dreamerv2, yarats2021drqv2} and use an action repeat of 2 for all tasks. Compared to the DMControl 100k benchmark shown in Table \ref{tab:planet}, we here consider more difficult tasks with up to $30\times$ more data. Results for DrQ-v2 and Dreamer-v2 are obtained from \citet{yarats2021drqv2, hafner2020dreamerv2}, results for DrQ are partially obtained from \citet{kostrikov2020image}, and results for CURL are reproduced using their publicly available implementation \citep{srinivas2020curl}. While baselines use task-dependent hyperparameters, TD-MPC uses the \textbf{same} hyperparameters for \textbf{all} tasks. Mean of 5 runs; shaded areas are 95\% confidence intervals. TD-MPC consistently outperforms CURL and DrQ, and is competitive with DrQ-v2 and Dreamer-v2.}
    \label{fig:dreamer}
    \vspace{-0.15in}
\end{figure*}

%% file: tables/walltime.tex
\begin{table}[t]
\vspace{-0.05in}
\caption{\textbf{Wall-time.} \textit{(top)} time to solve, and \textit{(bottom)} time per 500k environment steps (in hours) for the \textit{Walker Walk} and \textit{Humanoid Stand} tasks from DMControl. We consider the tasks solved when a method achieves an average return of 940 and 800, respectively. TD-MPC solves Walker Walk $\mathbf{16\times}$ faster than LOOP while using $\mathbf{3.3\times}$ less compute per 500k steps. Mean of 5 runs.}
\label{tab:walltime}
\vspace{0.075in}
\centering
\resizebox{0.48\textwidth}{!}{%
\begin{tabular}{lcccc|cc}
\multicolumn{1}{c}{} & \multicolumn{4}{c}{Walker Walk} & \multicolumn{2}{c}{Humanoid Stand} \\
\toprule
\textit{Wall-time} (h) & SAC & LOOP & MPC:sim & \textbf{TD-MPC} & SAC & \textbf{TD-MPC} \\\midrule
time to solve $\downarrow$     & $0.41$ & $7.72$ & $0.91$ & $0.47$ & $9.31$ & $9.39$ \\
h/500k steps $\downarrow$      & $1.41$ & $18.5$ & $-$ & $5.60$ & $1.82$ & $12.94$ \\
\bottomrule
\end{tabular}
}
\vspace{-0.2in}
\end{table}

%% file: tables/related-work.tex
\begin{table*}[t]
\caption{\textbf{Comparison to prior work.} We compare key components of \textbf{TD-MPC} to prior model-based and model-free approaches. \textit{Model objective} describes which objective is used to learn a (latent) dynamics model, \textit{value} denotes whether a value function is learned, \textit{inference} provides a simplified view of action selection at inference time, \textit{continuous} denotes whether an algorithm supports continuous action spaces, and \textit{compute} is a holistic estimate of the relative computational cost of methods during training and inference. We use \textit{policy w/ CEM} to indicate inference based primarily on a learned policy, and vice-versa.}
\label{tab:related-work}
\vspace{0.05in}
\centering
\resizebox{0.7\textwidth}{!}{%
\begin{tabular}{lccccc}
\toprule
\textbf{Method} & Model objective & Value & Inference & Continuous & Compute \\\midrule
SAC     & \color{BrickRed}{\XSolidBrush}      & \color{OliveGreen}{\Checkmark}      & \color{BrickRed}{Policy}        & \color{OliveGreen}{\Checkmark}     & \color{OliveGreen}{Low} \\
\colorcell QT-Opt     & \colorcell \color{BrickRed}{\XSolidBrush}      & \colorcell \color{OliveGreen}{\Checkmark}      & \colorcell \color{OliveGreen}{CEM}        & \colorcell \color{OliveGreen}{\Checkmark}     & \colorcell \color{OliveGreen}{Low} \\
MPC:sim     & \color{BrickRed}{Ground-truth model}      & \color{BrickRed}{\XSolidBrush}      & \color{OliveGreen}{CEM}        & \color{OliveGreen}{\Checkmark}     & \color{BrickRed}{High} \\
\colorcell POLO     & \colorcell \color{BrickRed}{Ground-truth model}      & \colorcell \color{OliveGreen}{\Checkmark}      & \colorcell \color{OliveGreen}{CEM}        & \colorcell \color{OliveGreen}{\Checkmark}     & \colorcell \color{BrickRed}{High} \\
LOOP     & \color{YellowOrange}{State prediction}      & \color{OliveGreen}{\Checkmark}      & \color{OliveGreen}{Policy w/ CEM}        & \color{OliveGreen}{\Checkmark}     & \color{YellowOrange}{Moderate} \\
\colorcell PlaNet     & \colorcell \color{YellowOrange}{Image prediction}      & \colorcell \color{BrickRed}{\XSolidBrush}      & \colorcell \color{OliveGreen}{CEM}        & \colorcell \color{OliveGreen}{\Checkmark}     & \colorcell \color{BrickRed}{High} \\
Dreamer     & \color{YellowOrange}{Image prediction}      & \color{OliveGreen}{\Checkmark}      & \color{BrickRed}{Policy}        & \color{OliveGreen}{\Checkmark}     & \color{YellowOrange}{Moderate} \\
\colorcell MuZero     & \colorcell \color{OliveGreen}{Reward/value pred.}      & \colorcell \color{OliveGreen}{\Checkmark}      & \colorcell \color{YellowOrange}{MCTS w/ policy}        & \colorcell \color{BrickRed}{\XSolidBrush}     & \colorcell \color{YellowOrange}{Moderate} \\
EfficientZero     & \color{OliveGreen}{Reward/value pred. + contrast.}      & \color{OliveGreen}{\Checkmark}      & \color{YellowOrange}{MCTS w/ policy}        & \color{BrickRed}{\XSolidBrush}     & \color{YellowOrange}{Moderate} \\
\midrule
\hcolorcell \textbf{TD-MPC} (ours)     & \hcolorcell \color{OliveGreen}{Reward/value pred. + latent pred.}      & \hcolorcell \color{OliveGreen}{\Checkmark}      & \hcolorcell \color{OliveGreen}{CEM w/ policy}        & \hcolorcell \color{OliveGreen}{\Checkmark}      & \hcolorcell \color{OliveGreen}{Low} \\
\bottomrule
\end{tabular}
}
\end{table*}